\documentclass{article}
\usepackage{arxiv}

\usepackage[utf8]{inputenc} 
\usepackage[T1]{fontenc}    
\usepackage{url}            
\usepackage{booktabs}       
\usepackage{amsfonts}       
\usepackage{nicefrac}       
\usepackage{microtype}      
\usepackage{lipsum}		
\usepackage{graphicx}
\usepackage{tabularx}
\usepackage[hidelinks]{hyperref}
\usepackage{multirow}
\usepackage{doi}
\usepackage{amsmath}
\usepackage{caption}
\usepackage[numbers,sort&compress]{natbib}

\newcounter{subfigpanel}[figure]

\newcommand{\subfiglabel}[1]{\refstepcounter{subfigpanel}\label{#1}}

\title{Visual Semantic Decoding of Electrocorticography from Video Stimuli using End-to-End Deep Learning}

\date{} 					

\author{
Stella Ho \\
Department of Biomedical Engineering \\
and Graeme Clark Institute\\
The University of Melbourne\\
Victoria, Australia\\
\And
Joel Villalobos \\
Department of Biomedical Engineering \\
and Graeme Clark Institute\\
The University of Melbourne\\
Victoria, Australia\\
\And
Joseph West \\
School of Computing and Information Systems \\
The University of Melbourne\\
Victoria, Australia\\
\And
Jingyang Liu \\
Bionics Institute \\
Victoria, Australia\\
\And
Weijie Qi \\
Department of Biomedical Engineering \\
and Graeme Clark Institute\\
The University of Melbourne\\
Victoria, Australia\\
\And
Haruhiko Kishima \\
Department of Neurosurgery \\
Graduate School of Medicine\\
The University of Osaka\\
Osaka, Japan\\
\And
Ryohei Fukuma \\
Department of Neuroinformatics \\
Graduate School of Medicine\\
The University of Osaka\\
Osaka, Japan\\
\And
Takufumi Yanagisawa \\
Department of Neuroinformatics \\
Graduate School of Medicine\\
The University of Osaka\\
Osaka, Japan\\
\And
Sam E.~John \\
Department of Biomedical Engineering \\
and Graeme Clark Institute\\
The University of Melbourne\\
Victoria, Australia\\
\And
David B.~Grayden\thanks{Corresponding author:~\texttt{grayden@unimelb.edu.au}} \\
Department of Biomedical Engineering \\
and Graeme Clark Institute\\
The University of Melbourne\\
Victoria, Australia\\
}

\date{}

\renewcommand{\headeright}{}
\renewcommand{\undertitle}{}
\renewcommand{\shorttitle}{}

\hypersetup{
pdftitle={A template for the arxiv style},
pdfsubject={q-bio.NC, q-bio.QM},
pdfauthor={David S.~Hippocampus, Elias D.~Striatum},
pdfkeywords={First keyword, Second keyword, More},
}

\begin{document}
\maketitle

\begin{abstract}
	ECoG-based visual semantic decoding enables inference of semantic interpretation of visual perception from complex, noisy brain activity. This study examines the feasibility of visual semantic decoding using an end-to-end deep learning framework using electrocorticography (ECoG). Specifically, the decoding task is to predict visual categories from video stimuli using time-series neural inputs. A previously collected ECoG dataset from participants ($n=17$) with drug-resistant epilepsy is used for analysis. With fewer than 50 training samples per visual category, this study evaluates multiple deep learning approaches, artificial neural network architectures, and frequency-band filtered inputs. The best-performing approach is analyzed to shed light on the discriminative information it relies on across spectral, temporal, and cortical dimensions. The selected decoding system uses mixup augmentation, a Transformer-based encoder, and high-gamma (80-150~Hz) inputs with a 900~ms post-stimulus window. Further analysis shows that early visual cortex (V2--V4), ventral stream visual cortex, MT+ complex with neighbouring visual areas, and lateral temporal cortex contributed substantially to decoding performance. This study demonstrates that an end-to-end deep learning framework can yield promising decoding performance from dynamic visual stimuli without handcrafted features, while the model behavior remains interpretable through spectral, temporal, and cortical dimensions, which are broadly consistent with established neuroscience knowledge.
\end{abstract}


\section{Introduction}
The human cortex perceives, interprets, and acts upon incoming sensory information. Neural decoding provides a computational framework for extracting and interpreting this information from signals recorded from the brain by electroencephalography (EEG) or other recording methods. In sensory neuroscience, neural decoding is used to study how sensory input is represented in brain activity and how these neural representations reflect perception \citep{holdgraf2017encoding}. Machine learning provides a way to learn mappings between stimulus-evoked brain activity and relevant variables \citep{glaser2020machine,livezey2021deep}, such as stimulus labels. In this bottom-up decoding process, the mapping between neural population responses and stimulus information is inherently non-linear \citep{yang2021revealing}. To model this nonlinearity, deep learning (DL) is one of the most suitable approaches \citep{livezey2021deep}. An end-to-end DL framework allows for a direct mapping of neural inputs to high-level information outputs without handcrafted features or fixed priors \citep{schirrmeister2017deep}.

A DL end-to-end approach brings both opportunities and challenges, as electrophysiological signals are inherently complex, noisy, and dynamic compared to image processing in computer vision and text analysis in natural language processing. This imposes stringent requirements on the quality and quantity of the data for the machine learning process. In particular, clinical settings often introduce heterogeneity into the data. For example, clinical data are often collected from subjects with severe neurological conditions, such as epilepsy, which allows for invasive approaches. Subdural EEG electrodes are implanted for individual medical reasons, so the resulting electrode coverage and density, as well as individual response patterns, vary significantly across subjects. As a result, decoding models are often highly personalised. The resulting limited availability of subject-specific data poses a major hurdle for training data-hungry end-to-end DL frameworks.

Model interpretability presents another significant challenge since DL models are often described as black boxes so their decision-making process can be opaque and difficult to interpret \citep{rudin2019stop}. In clinical brain-computer interface (BCI) settings, this is a non-negligible concern, as the output from DL models may be unreliable or even lead to unsafe decisions. For example, the model may rely on signal components, such as artifacts or other confounding factors, that are correlated with the labels but not necessarily related to the target neural process, while still showing good performance on the test set \citep{west2023machine}. Hence, in neuroscience and clinical applications, it is important to examine whether the model behavior is consistent with domain knowledge, such as expected cortical responses, instead of merely relying on performance metrics, such as accuracy, F1-score, and area under the receiver operating characteristic curve (AUC-ROC) score.

Beyond these DL challenges, neural decoding using dynamic visual stimuli introduces another challenge. Real-world visual perception is dynamic and continuous, while visual decoding has commonly been studied using static images as sensory stimuli \citep{kay2008identifying, majima2014decoding, rupp2017semantic, gifford2022large, guenther2024image, ferrante2024decoding}. In visual neural decoding, video stimuli have attracted increasing interest in neuroscience studies \citep{nishimoto2011reconstructing, huth2016decoding, isik2018what, berezutskaya2020cortical} as they are more naturalistic and offer richer context. However, it remains relatively under-explored compared to static image paradigms since it introduces a series of challenges, such as lack of well-defined stimulus onset \citep{isik2018what} and confounded visual information (e.g., object and action) \citep{huth2016decoding}.

Motivated by these challenges, this study focuses on visual semantic decoding using an end-to-end framework to predict visual semantic categories from time-series neural signals. The dataset used was previously collected by \citep{fukuma2022voluntary}, and contains electrocorticography (ECoG) recordings from a clinical cohort of participants ($n=17$) with drug-resistant epilepsy during a video-watching task. In this work, the goal is to establish a practical and reliable approach for visual semantic decoding using simple and effective DL architectures and approaches, rather than developing a novel architecture. 
 
The main contributions are as follows:
\begin{enumerate}
    \item To address the limited data of the clinical intracranial recordings, we evaluate three well-established DL approaches: few-shot learning via prototypical networks \citep{DBLP:conf/nips/SnellSZ17}, data augmentation via mixup \citep{DBLP:conf/icml/VermaLBNMLB19}, and contrastive representation learning via SimCLR \citep{DBLP:conf/icml/ChenK0H20}.

    \item We compare multiple artificial neural network architectures and frequency-band filtered signals (i.e., alpha, beta, low-gamma, high-gamma, broadband-gamma, broadband) and identify the Transformer as the best-performing encoder and high-gamma (80-150~Hz) as the most informative time-series input.

    \item We observe that the Transformer-based model has limited spectral interpretability when directly processing broadband inputs, and that high-gamma spectral power is a major contributor to its decoding performance.

    \item We analyse post-stimulus window size and cortical region importance, showing that 900~ms is the group-level optimal window, and that early visual cortex (V2--V4), ventral stream visual cortex, MT+ complex with neighbouring visual areas, and lateral temporal cortex reflect strong individual contributions to model inference.

    \item We demonstrate that the behavior of the best-performing model can be interpreted using neurobiological domain knowledge, thereby showing it to be more reliable.   
\end{enumerate}

\section{Methodology}
\subsection{Analyzed Dataset}
We used the ECoG dataset of a video perception task that was originally collected and analyzed by \citep{fukuma2022voluntary} on patients undergoing assessment of drug-resistant epilepsy. This dataset includes 17 participants (E01–E17) implanted with subdural electrodes covering the occipital and/or temporal lobes for clinical reasons. The ECoG recordings were collected over six separate sessions. In each session, the participants watched a 10-minute video. These videos were created from movie trailers, behind-the-scenes videos, and animation clips, which cover a wide range of visual semantic content. During the experiment, ECoG signals were recorded at 10~kHz by EEG-1200, a clinical EEG recording system (Nihon Kohden, Tokyo, Japan) \citep{fukuma2022voluntary}. The original data collection was approved by the ethics committees of the participating hospitals, including Osaka University Medical Hospital (Approval No. 14353, UMIN000017900), Juntendo University Hospital (Approval No. 18-164), and Nara Medical University Hospital (Approval No. 2098). Written consent was obtained from all participants

Similarly to \citep{fukuma2022voluntary}, we focused on three distinct visual semantic categories identified in their benchmark: Human Face, Text, and Landscape. We curated a dataset tailored for label-scarce classification benchmarks, consisting of 150 labeled and 3,450 unlabeled non-overlapping ECoG segments. Specifically, to ensure label quality, we manually selected the samples in which the target category was semantically dominant. The dataset was restricted to 50 examples per visual category. For Leave-One-Recording-Out (LORO) evaluation, all labeled ECoG samples from one complete recording session were held out as the test set in each fold, while labeled samples from the remaining five sessions formed the training/validation set. The number of labeled training and test samples in each LORO fold is detailed in Table \ref{tab:data_info}. 

\begin{table}[htbp]
\centering
\caption{Data distribution for the Leave-One-Recording-Out (LORO) cross-validation. For each fold, one recording is held out for testing, while the remaining recordings are the training set.}
\label{tab:data_info}
\small
\begin{tabular}{l cc cc cc cc}
\toprule
\multirow{2}{*}{\textbf{Held-out}} & \multicolumn{4}{c}{\textbf{Train}} & \multicolumn{4}{c}{\textbf{Test}}\\
\cmidrule(lr){2-5}  \cmidrule(lr){6-9} 
& Human Face & Text & Landscape & Overall & Human Face & Text & Landscape & Overall  \\
\midrule
Trn01 & 40 & 37 & 40 & 117 & 10 & 13 & 10 & 33 \\
Trn02 & 42 & 40	& 44 & 126 & 8  & 10 & 6 & 24 \\
Trn03 & 45	&46	&44 & 135 &5 &4	&6 & 15 \\
Trn04 & 44	&37	&40 & 121 &6 &13 &10 & 29 \\
Trn05 & 41	&45	&41 & 127 &9 &5	&9 & 23 \\
Trn06 & 38	&45	&41 & 124 &12	&5	&9 & 26 \\
\bottomrule
\end{tabular}
\end{table}

\subsubsection{Signal Preprocessing}  The raw signals were preprocessed using a notch filter at the corresponding power-line frequency (50 or 60~Hz, depending on the recording site) and a 5th-order high-pass Butterworth filter with a cutoff frequency of 3~Hz. The signals were then downsampled to 500~Hz using polyphase resampling. Subsequently, the ECoG recordings were segmented into non-overlapping 500~ms windows, each spanning from stimulus onset to 500~ms post-onset, as previous results showed that stimulus categories can be decoded from ECoG epochs within the first 500~ms after stimulus onset~\citep{kapeller2018realtime}.

\subsubsection{Electrode Selection} We discarded channels in regions that were unlikely to be functionally related to visual/semantic processing, including sensory-motor, auditory, and non-task-relevant frontal areas. The cortical regions retained for analysis included Primary Visual Cortex (V1), Early Visual Cortex (V2--V4), Dorsal Stream Visual Cortex, Ventral Stream Visual Cortex, MT+ Complex / neighboring visual areas, Medial Temporal Cortex, Lateral Temporal Cortex, Temporo-Parieto-Occipital Junction, Superior Parietal Cortex, Inferior Parietal Cortex, Posterior Cingulate Cortex, Inferior Frontal Cortex, and Dorsolateral Prefrontal Cortex.

\subsection{Deep Learning Approach}
To address the inherent challenge of label scarcity in clinical BCI applications, we assessed three distinct and representative low-resource deep learning approaches. We evaluated their ability to extract robust features from high-dimensional noisy neural signals.

\subsubsection{Few-shot Learning via Prototypical Networks} Prototypical Networks (ProtoNet) \citep{DBLP:conf/nips/SnellSZ17} employ an episodic training framework. In this experiment, ProtoNet was evaluated under a 3-way 5-shot classification setting. Specifically, in each episode, the model maps 5 support samples per class into an embedding space and averages their embeddings to generate a class-specific prototype for each visual category. Query samples are then classified based on their squared Euclidean distance to these prototypes. This method allows the model to learn geometric similarity in the latent space rather than relying on fixed decision boundaries.

\subsubsection{Data Augmentation via Mixup} 
This multi-level Mixup method \citep{DBLP:conf/icml/VermaLBNMLB19} stochastically interpolates samples at either the input level or the first latent embedding level. Given two training ECoG samples $(\boldsymbol{x}_i, \boldsymbol{y}_i)$ and $(\boldsymbol{x}_j, \boldsymbol{y}_j)$ and a mixing coefficient $\lambda$, the augmented representation is computed as follows:
\begin{itemize}
    \item Input Mixup (Layer 0) \citep{DBLP:conf/iclr/ZhangCDL18}: Interpolation is performed directly on the raw ECoG signals, as $\tilde{\boldsymbol{x}} = \lambda \boldsymbol{x}_{i} + (1-\lambda)\boldsymbol{x}_{j}$.
    \item  Manifold Mixup (Layer 1) \citep{DBLP:conf/icml/VermaLBNMLB19}: Interpolation is applied on the hidden state of the first latent layer, as $\tilde{\boldsymbol{h}} = \lambda \boldsymbol{h}_{i} + (1-\lambda) \boldsymbol{h}_{j}$.
\end{itemize}
The resulting augmented representations propagate through subsequent layers to output a prediction $\hat{\boldsymbol{y}}$. The model is then optimised using a weighted cross-entropy loss $\mathcal{L}_\mathrm{CE}$, defined as $\mathcal{L}_\mathrm{mixup} = \lambda \mathcal{L}_\mathrm{CE}(\hat{\boldsymbol{y}},\boldsymbol{y}_i)+(1-\lambda)\mathcal{L}_{\mathrm{CE}}(\hat{\boldsymbol{y}},\boldsymbol{y}_j)$. By creating a continuous representation manifold, this approach allows the model to learn the transitions between examples, resulting in smooth decision boundaries.

\subsubsection{Self-Supervised Contrastive Learning} 
Self-supervised representation learning is tailored to a specific clinical BCI scenario where unlabeled ECoG recordings are abundant but expert annotation is expensive. To leverage the unlabeled data pool, we used the Self-Supervised Contrastive Learning (SimCLR) framework \citep{DBLP:conf/icml/ChenK0H20} for representation learning. There are two stages in this approach:
\begin{itemize}
    \item Pretraining: Two augmented views of each input ECoG epoch are created by applying channel dropout (0.2) \citep{DBLP:conf/icassp/SaeedGPZ21,JMLR:v15:srivastava14a}, random time shifts ($\pm20$ ms) and Gaussian noise ($\sigma=0.01$). These views pass through a shared encoder and pooling module into a projection head. The model is trained using contrastive loss, where the objective is to map different variations of the same sample closer together while pushing them away from other negative samples in the batch.

    \item Downstream classification: The projection head is discarded and the pre-trained encoder weights are used but frozen during training. The subsequent blocks, including the pooling, two-layer MLP projection and classifier head, are then trained via cross-entropy loss using the labeled dataset.
\end{itemize} 
By learning discriminative embeddings, this method bypasses reconstruction approaches (e.g., autoencoder) that are computationally expensive and unsuitable for noisy, complex high-dimensional signals, such as ECoG. Self-supervised SimCLR pretraining was performed once per subject using all available unlabelled ECoG samples. The test-set were held out from each LORO fold and the downstream supervised classifier. 

Although unlabeled epochs from the same recording as the test-set were used in pretraining, this is not purely \textit{transductive pre-training} \citep{DBLP:conf/emnlp/OuchiSI19} because the labeled test epochs themselves were excluded and segmentation used non-overlapping windows, this prevents test-instance exposure during representation learning. Prior works, do however indicate possible leakage when using adjacent neurological signals \citep{west2023machine} as the representation may still have been exposed to recording-level or session-level structure from the recording that the test-set was extracted. Considering possible adjacency leakage, our approach sits between transductive \citep{DBLP:conf/emnlp/OuchiSI19, bianco2016curl} and inductive pre-training \citep{bianco2016curl} and remains valid.

\subsection{Model Architecture}
\begin{figure}[htbp]
 \centering
        \includegraphics[width=0.85\textwidth]{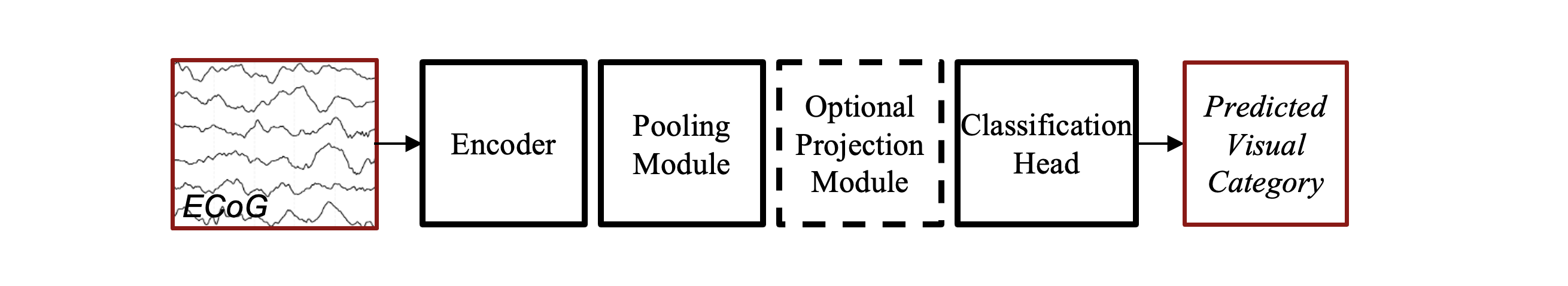}
        \vspace{-10pt}
 \caption{Schematic of the end-to-end neural decoding framework. Note that this architecture did not apply to the Prototypical Network, where we replaced the classifier head with a distance-based similarity mechanism.}
\label{fig1:schematic}
\end{figure}

The neural decoding framework consists of four main components (see Figure \ref{fig1:schematic}): (1) an encoder that extracts artificial neural representations from multi-channel time-series inputs; (2) a temporal pooling module that aggregates the encoder output into a fixed-size vector, implemented as either global average pooling or learnable query-key temporal attention pooling \citep{DBLP:conf/icml/IlseTW18,DBLP:conf/nips/VaswaniSPUJGKP17}; (3) an optional projection head (two-layer MLP with ReLU activations) that further transforms the encoded embedding to a lower-dimensional vector before classification; and (4) a classification head (a fully connected linear layer) that maps the neural representation to visual categories. The specific configuration of pooling method and projection head for each encoder (see Table \ref{tab:config} in the Supplementary Material) was determined based on preliminary studies. Due to different input dimensions across subjects, the hyperparameters were tuned using the same Bayesian optimization protocol on the training set before model training and evaluation.

\subsubsection{Encoder}
We examined six lightweight yet effective encoders, each with a known ability to capture temporal dependencies in time-series data: 

\begin{itemize}
    \item \textbf{Residual Network (ResNet)} \citep{DBLP:conf/cvpr/HeZRS16}: a 1D Convolutional Neural Network (1D CNN) that processes signals in the temporal domain. It consists of stacked residual blocks \citep{DBLP:conf/cvpr/HeZRS16} with a pre-activation design \citep{DBLP:conf/eccv/HeZRS16} in which batch normalisation and activation are applied before each convolution. 

    \item \textbf{Large-Small Network (LSNet)}: a 1D CNN with two parallel convolutions of different kernel sizes adapted from \citep{DBLP:conf/cvpr/WangCLHD25}. In each block, a larger kernel captures broader temporal context, while a smaller kernel captures local details. The resulting outputs from both convolutions are concatenated to form a multi-scale neural representation.

    \item \textbf{Temporal Convolutional Network (TCN)}: a 1D CNN that applies dilated convolutions in the temporal domain \citep{Bai2018AnEE}. In this study, we used symmetric (non-causal) padding to leverage the full temporal context.

    \item \textbf{EEGNet \citep{lawhern2018eegnet}}: a CNN-based neural network for EEG-based BCIs, consisting of temporal convolution, depthwise spatial convolution, and separable convolution layers. In this study, we adapted EEGNet encoder to our ECoG decoding pipeline.

    \item \textbf{Bidirectional Long Short-Term Memory (BiLSTM)} \citep{DBLP:journals/nn/GravesS05}: a recurrent neural network that processes signals in both forward and backward directions using gated recurrence and memory cells.

    \item \textbf{Transformer}: a Transformer encoder that uses a pointwise 1D convolutional layer to project the input channel into a fixed hidden dimension, followed by stacked multi-head self-attention blocks \citep{DBLP:conf/nips/VaswaniSPUJGKP17} operating across the temporal dimension.
    
\end{itemize}

\subsection{Evaluation Protocol} \label{subsec:eval}
Data leakage through temporally related segments and noise complexity has been noted as a significant confound resulting in high reported performance but low clinical relevance \citep{west2023machine}. We carefully mitigate this concern by temporally separating the test samples from the training samples. Performance was assessed using LORO cross-validation across six different sessions, where all samples from one complete session was held out as the test set in each fold. This protocol was designed to mitigate the impact of factors that may vary across sessions, such as subject fatigue or task unfamiliarity. To account for stochastic initialisation in artificial neural network training, each fold was evaluated with 10 random seeds.

To address class imbalance, we report both per-class and balanced accuracy. The Balanced Accuracy ($BA$) was calculated as
\begin{equation}
BA = \frac{1}{N_c} \sum_{c}^{N_c} \frac{\mathrm{TP}_{c}}{\mathrm{TP}_{c}+\mathrm{FN}_{c}}
\end{equation}
where $\mathrm{TP}_{c}$ and $\mathrm{FP}_{c}$ represent the true positives and false positives for class $c$, respectively. Model performance for each subject is reported as the mean and standard deviation (SD). The mean value represents the average across 6 held-out recordings, which are each averaged over 10 random seeds, while the standard deviation indicates the performance variability across the 6 different recording sessions.

We also report Cohen's Kappa ($\kappa$) \citep{Cohen1960ACO}, which evaluates classification performance relative to chance level and is defined as 
\begin{equation}
\kappa = \frac{p_o - p_e}{1 - p_e}
\end{equation}
where $p_o$ is the proportion of correct predictions and $p_e$ is the proportion expected by random guessing. We report $BA$ as our primary evaluation metric and $\kappa$ as a supplementary measure.

\subsubsection{Significance Test} For model evaluation at the individual level, we carried out two-tailed paired t-test to assess statistical significance. Performances were paired by test session and random seed ($n=60$) to address inter-session variance and initialisation bias. For the cohort-level analysis study ($n=17$), we applied the Wilcoxon Signed-Rank Test to account for individual differences. Bonferroni correction was applied to control the chance of false positives across multiple comparisons.

\subsubsection{Chance Level} We performed a training-label scramble analysis to estimate chance-level BA. Across individual subjects and frequency bands, the estimated empirical chance level ranged from 0.313 to 0.384, with a mean of 0.337, which supports the theoretical chance level of 0.333. Therefore, 0.333 is used as the chance level in this study.

\section{Results}
In this study, we first conducted a proof-of-concept analysis on a representative single subject to validate the feasibility of the end-to-end framework and to identify a robust learning approach and model architecture for visual semantic decoding. This allowed us to bypass the need for an exhaustive and computationally expensive search. This staged selection procedure was used to identify a practical decoding system for subsequent cohort-level analysis. The selected decoding system was subsequently applied to the full cohort of 17 participants, where we focused on investigating informative frequency bands for visual semantic decoding.

\subsection{Proof-of-Concept: End-to-End Visual Decoding}
We selected Subject E02 as the representative case as the electrode array of this participant covered several key visual processing regions, including the primary and early visual cortices (V1--V4), the dorsal and ventral visual streams, and the MT+ complex \citep{fukuma2022voluntary}.

\begin{table*}[htbp]
\centering
\caption{Performances of different learning approaches for Subject E02 ($n=1$). This experiment was conducted on broadband ECoG signals using the ResNet encoder with global averaging pooling and an MLP projection head as the benchmark architecture. The projection head was included to ensure a fair comparison across learning approaches. \underline{Underline} indicates the best performance.}
\label{tab:exp_1}
\small
\begin{tabular}{l cc cc cc cc c}
\toprule
\multirow{2}{*}{\textbf{Method}} & \multicolumn{2}{c}{\textbf{Human Face}} & \multicolumn{2}{c}{\textbf{Text}} & \multicolumn{2}{c}{\textbf{Landscape}} & \multicolumn{2}{c}{$\mathbf{BA}$} &\textbf{Kappa}\\
\cmidrule(lr){2-3} \cmidrule(lr){4-5} \cmidrule(lr){6-7} \cmidrule(lr){8-9} 
& Mean & SD & Mean & SD & Mean & SD & Mean & SD & $\kappa$ \\
\midrule
Baseline & 0.459 & 0.13 & 0.515 & 0.12 & 0.330 & 0.07 & 0.435 & 0.08 &0.151 \\
ProtoNet & 0.488 & 0.14 & 0.583 & 0.07 & 0.480 & 0.06 & 0.517 & 0.07 & 0.271\\ 
Mixup & 0.520 & 0.09 & 0.659 & 0.17 & 0.386 & 0.08 & 0.522 & 0.08 &0.269\\
SimCLR  & 0.513 & 0.15 & 0.681 & 0.16 & 0.500 & 0.15 & \underline{0.565} & 0.10 & 0.335\\
\bottomrule
\end{tabular}
\end{table*}

\subsubsection{Deep Learning Approach}
Table \ref{tab:exp_1} summarises the performances of four different learning approaches. The baseline used standard supervised training with cross-entropy loss. The result shows that all three learning frameworks significantly outperformed the baseline ($p<0.001$). SimCLR achieved the highest overall $BA$ of 0.565 and kappa $\kappa$~=~0.335, followed by Mixup, which yielded a $BA$ of 0.522 and $\kappa$~=~0.269. Values $\kappa >$~0.25 indicate that the performance was above chance level, suggesting that the model extracted task-relevant neural information from ECoG. In addition, the ``Text" class showed consistently higher accuracy than the ``Human Face" and ``Landscape" classes. This was likely due to strong and consistent neural responses in the early visual cortex elicited by high-contrast low-level visual features from text stimuli, which were easier for the model to capture. 

Although SimCLR achieved the highest overall BA in this benchmark, Mixup was preferred as a more practical and computationally efficient approach. SimCLR requires an additional pretraining stage to learn a robust representation space from a large number of ECoG epochs, which increases computational overhead and training complexity. In some settings, adapting test-session samples to this representation space may require extra training, making it less suitable for low-latency, real-time BCI applications. Furthermore, its two-stage training protocol may trigger representation collapse within the feature space due to the model's shortcut learning \citep{DBLP:journals/natmi/GeirhosJMZBBW20}. In contrast, Mixup employs a single-stage supervised training approach and showed only a marginal accuracy degradation compared to SimCLR. Hence, Mixup was selected for the subsequent experiments. To support robust performance, each encoder was configured with a specific pooling method and projection setting (with or without projection) under Mixup augmentation. These configurations were determined through preliminary ablation studies and summarised in Table \ref{tab:config} in the Supplementary Material.

\begin{table*}[htbp]
\centering
\caption{Encoder performance on broadband (3-250~Hz) and high-gamma (80-150~Hz) signals with Mixup. The evaluation was conducted on Subject E02. \underline{Underlined} indicates the best performance.}
\label{tab:exp_2}
\small
\begin{tabular}{l cc cc cc cc c}
\toprule
\multirow{2}{*}{\textbf{Encoder}} & \multicolumn{2}{c}{\textbf{Human Face}} & \multicolumn{2}{c}{\textbf{Text}} & \multicolumn{2}{c}{\textbf{Landscape}} & \multicolumn{2}{c}{$\mathbf{BA}$} & {\textbf{Kappa}}\\
\cmidrule(lr){2-3} \cmidrule(lr){4-5} \cmidrule(lr){6-7} \cmidrule(lr){8-9}
 & Mean & SD & Mean & SD & Mean & SD & Mean & SD & $\kappa$ \\
\midrule
\addlinespace
\multicolumn{5}{l}{\textbf{Mixup w/t. Broadband} (3-250Hz)}\\
\midrule
ResNet & 0.523 & 0.16 & 0.758 & 0.11 & 0.492 & 0.09 & 0.591 & 0.07 & 0.371 \\
LSNet & 0.595 & 0.12 & 0.628 & 0.13 & 0.402 & 0.06 & 0.542 & 0.07 & 0.300\\
TCN & 0.452 & 0.09 & 0.700 & 0.18 & 0.392 & 0.11 & 0.515 & 0.07 & 0.251 \\
EEGNet & 0.495 & 0.20 & 0.712 & 0.17 & 0.446 & 0.12 & 0.551 & 0.12 & 0.317 \\
BiLSTM & 0.579 & 0.23 & 0.703 & 0.15 & 0.564 & 0.12 & \underline{0.615} & 0.09 & 0.410\\
Transformer & 0.481 & 0.19 & 0.760 & 0.17 & 0.449 & 0.05 & 0.563 & 0.08 & 0.323 \\
\midrule
\addlinespace
\multicolumn{6}{l}{\textbf{Mixup w/t High-gamma} (80-150Hz)}\\
\midrule
ResNet & 0.678 & 0.07 & 0.552 & 0.05 & 0.402 & 0.09 & 0.544 & 0.02 & 0.301\\
LSNet & 0.685 & 0.06 & 0.652 & 0.14 & 0.363 & 0.11 & 0.566 & 0.04 &  0.325\\
TCN & 0.823 & 0.10 & 0.544 & 0.07 & 0.108 & 0.04 & 0.492 & 0.04 &  0.230 \\
EEGNet & 0.597 & 0.13 & 0.608 & 0.07 & 0.550 & 0.12 & 0.585 & 0.06 & 0.359 \\ 
BiLSTM & 0.612 & 0.13 & 0.657 & 0.16 & 0.641 & 0.13 & 0.637 & 0.05 & 0.435\\
Transformer & 0.704 & 0.12 & 0.699 & 0.12 & 0.647 & 0.09 & \underline{0.683} &0.04 & 0.505\\
\bottomrule
\end{tabular}
\end{table*}

\subsubsection{Encoder Architecture} 
To narrow down the selection of encoder architectures, we evaluated candidate encoders on two frequency ranges: broadband (3-250~Hz) and high-gamma band (80-150~Hz). Broadband signals reflect neural activity across a wide frequency range. This allows the model to learn task-relevant information from multiple frequency-band components. High-gamma activity has been shown to reflect stimulus-related cortical responses \citep{Brunet2019HumanVC, Bartoli2019GammaVisualCortex}. As shown in Table \ref{tab:exp_2}, classification accuracy was highly dependent on both the encoder architecture and the input frequency band. Across all candidate architectures, the Transformer achieved the highest overall decoding performance with a $BA$ of 0.683 on the high-gamma band, followed by the BiLSTM at 0.637. The ResNet achieved a performance of 0.591 when using the broadband signal. Based on these results, we selected three top-performing encoders: the Transformer, the BiLSTM, and the ResNet for subsequent analysis.

\subsection{Cohort Study: Frequency-Band and Model Analysis} \label{subsec:cohort}
As the previous experiment focused on proof of concept with a single subject, we extended the spectral analysis to a greater number of subjects. We band-pass filtered the continuous ECoG signals using a 5th-order Butterworth filter into five frequency ranges (alpha: 8-13~Hz, beta: 13-30~Hz, low-gamma: 40-80~Hz, broadband-gamma: 40-150~Hz, and high-gamma: 80-150~Hz), resulting in separate time-series inputs of each band. The decoding performance was then evaluated across these inputs and the top-performing models for three subjects ($n=3$). The best decoding system was further assessed across the whole cohort ($n=17$), which included substantial differences in subdural electrode coverage.

\begin{figure*}[htbp]
    \centering
    \begin{minipage}{0.99\linewidth}
        \raggedright
        \textbf{A}\par\vspace{0.10cm}
        \includegraphics[width=\linewidth]{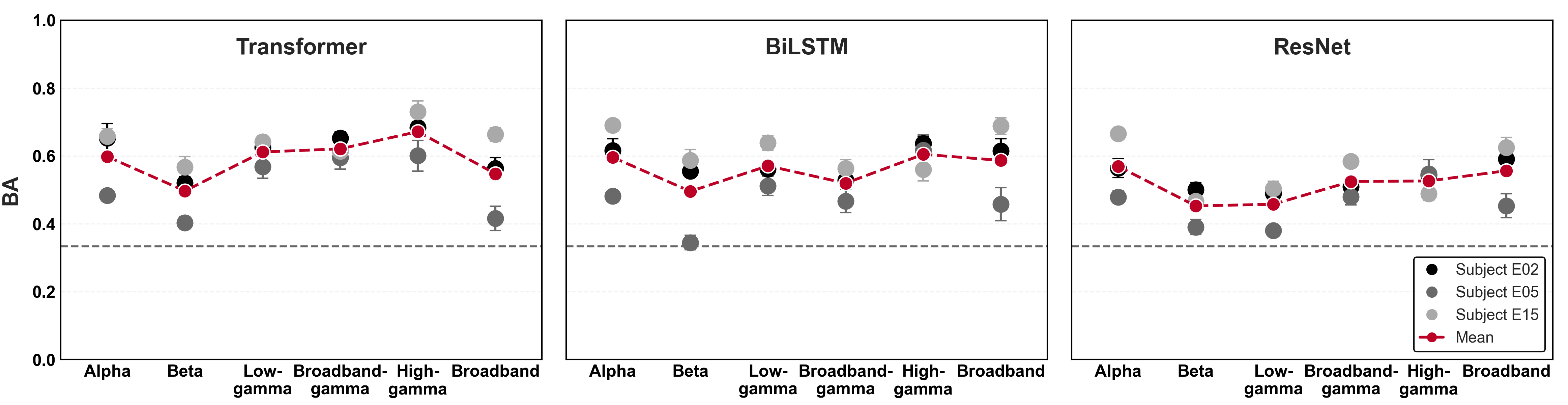}
    \end{minipage}
    \begin{minipage}[t]{0.43\linewidth}
        \raggedright
        \textbf{B}\par\vspace{0.10cm}
        \includegraphics[width=\linewidth]{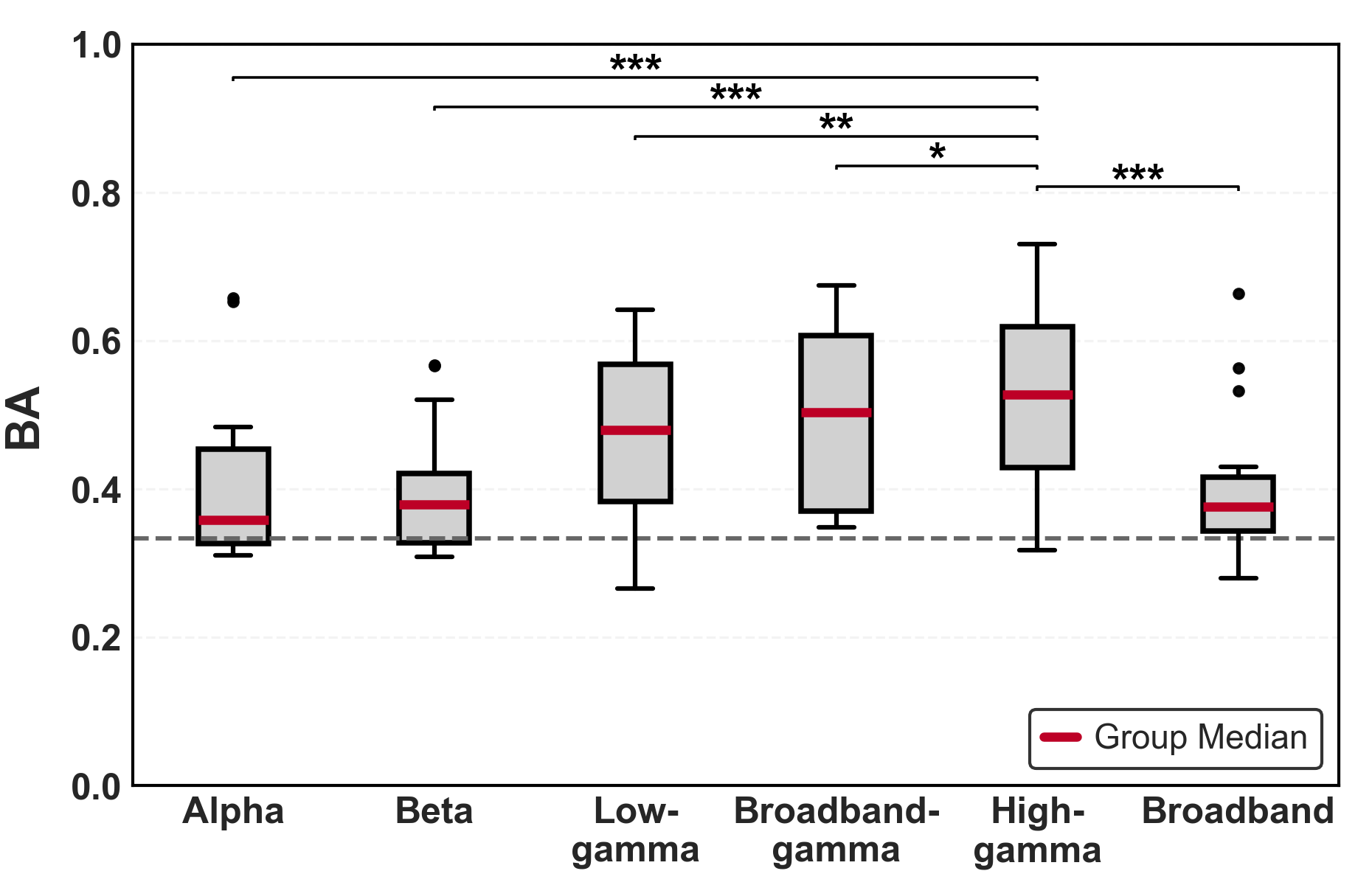}
    \end{minipage}
    \begin{minipage}[t]{0.55\linewidth}
        \raggedright
        \textbf{C}\par\vspace{0.10cm}
        \includegraphics[width=\linewidth]{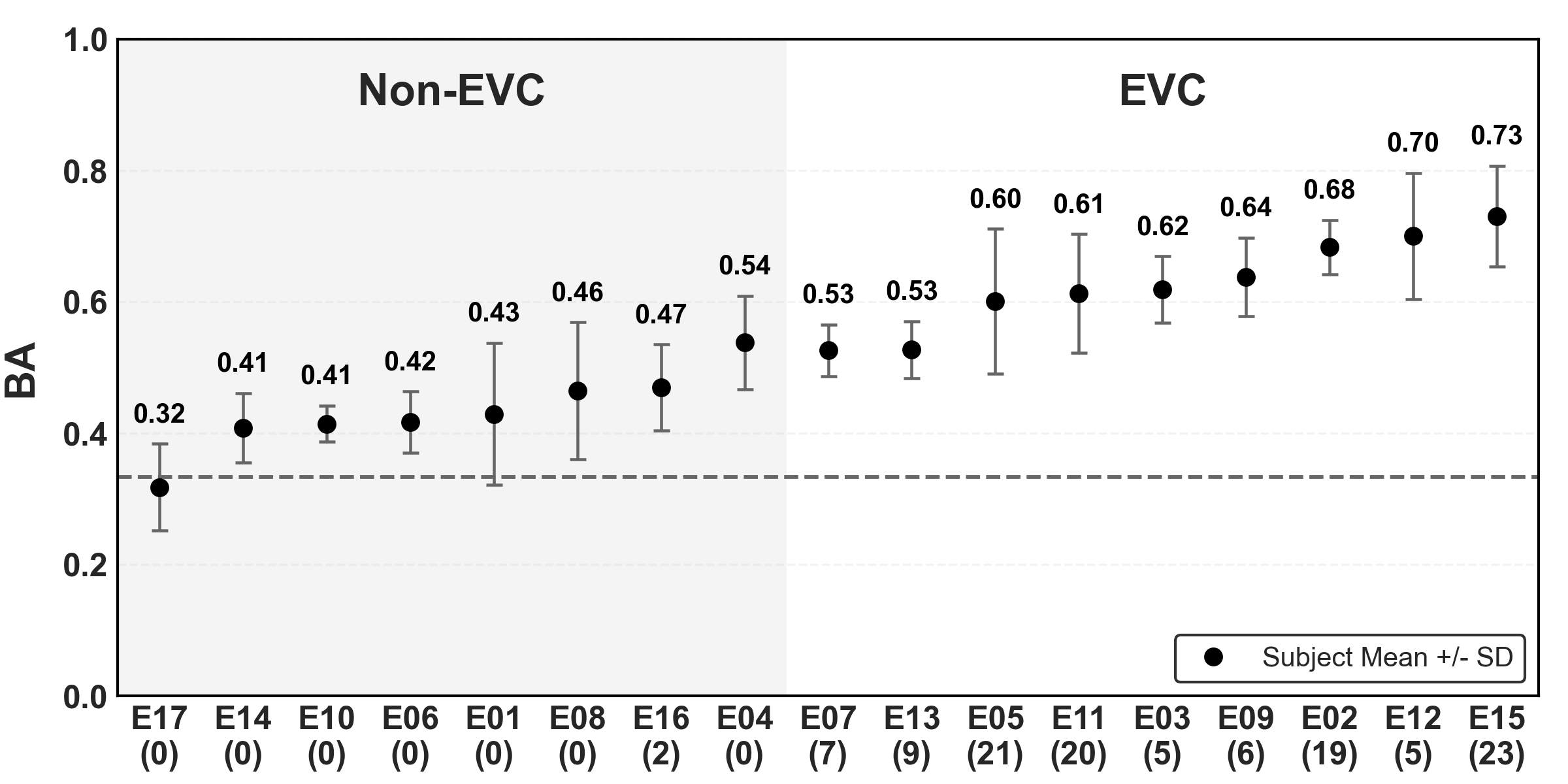}
        \label{fig:spec_subgroup}
    \end{minipage}
    \caption{\textbf{A}.~Decoding accuracy across frequency bands for three encoder architectures (Transformer, BiLSTM, ResNet) in three representative subjects ($n=3$) indicated by different dot shadings. Mean results are shown in red and connected by lines to aid visualization. \textbf{B}.~Decoding accuracy by frequency band using the Transformer encoder ($n=17$). Boxes show the interquartile range, red centre lines represent group medians, and black circles represent outliers. Asterisks indicate significance levels from pairwise Wilcoxon signed-rank tests after correction: $^{***}p<0.001$, $^{**}p<0.01$, $^{*}p<0.05$; n.s., not significant. \textbf{C}.~Decoding accuracy across all subjects using the Transformer encoder with high-gamma inputs. Points show subject-level means, error bars indicate standard deviation across test folds, and numbers above points indicate the mean values. Subjects are grouped as EVC or non-EVC according to whether they had ($\geq$ 5) electrodes across V1--V4. The numbers of subdural electrodes located in V1--V4 areas are shown in parentheses under subject IDs. Subjects are ranked within each group by mean BA.}
    \label{fig:spectral}
    \subfiglabel{fig:spec}
    \subfiglabel{fig:spec_all}
    \subfiglabel{fig:spec_subgroup}
\end{figure*}

\subsubsection{Model Selection $(n=3)$}\label{subsec:rep_sub}
To identify the optimal encoder architecture, we performed a targeted analysis of three subjects (E02, E05, E15). These participants were chosen as they had broad electrode coverage across their visual cortical areas, including primary visual cortex (V1), early visual cortex (V2--V4), dorsal-stream visual cortex, ventral-stream visual cortex, and the MT+ complex with neighbouring visual areas \citep{fukuma2022voluntary}.

Figure \ref{fig:spec} compares decoding accuracy across the frequency bands for three candidate encoders. The Transformer encoder achieved its highest group-level mean accuracy of 0.672 in the high-gamma band. This high-gamma accuracy exceeded the other frequency band inputs by 0.051 to 0.175, and the alpha band input also showed relatively high accuracy of 0.598. This performance was not consistently observed across the full cohort ($n=17$; see Figure \ref{fig:spec_all}). Similarly, the BiLSTM showed its highest mean $BA$ of 0.605 with high-gamma input. The ResNet yielded a high-gamma mean $BA$ of 0.526, with a slightly higher broadband mean $BA$ of 0.556, but it consistently underperformed the other two encoders. As a result, we selected the Transformer (detailed in Table \ref{tab:transformer} in the Supplementary Material) for the subsequent study.

\subsubsection{Frequency-Band Analysis ($n=17$)} 
As shown in Figure \ref{fig:spec_all} and Table \ref{tab:freq},  the high-gamma input showed strong performance, with the highest group-level median accuracy of 0.527 across the cohort ($n=17$). This suggests that high-gamma activity carries highly discriminative information for visual category decoding, which broadly aligns with previous findings \citep{fukuma2022voluntary,rupp2017semantic}. The broadband-gamma and low-gamma inputs achieved median accuracies of 0.503 and 0.479, respectively. These results highlight the discriminative power of gamma band activity. In comparison, the alpha and beta band inputs exhibited near-chance level (0.333) performance, with median accuracies of 0.358 and 0.379, respectively. The median accuracy for broadband input was 0.376, which remained significantly lower than high gamma by 0.151 ($p<0.001$).

\section{Analyses and Discussion}
\subsection{Impact of Early Visual Cortical Regions}
Figure \ref{fig:spec_all} shows inter-subject decoding variability, particularly in the gamma bands. Since the visual processing pathways are likely to be crucial for decoding, we asked the question: Is early visual cortical (EVC) region coverage a contributing factor to decoding performance? To study this, the cohort was partitioned into two groups: an EVC subgroup ($n = 9$), consisting of subjects with at least five electrodes within the EVC region (V1–V4), and a Non-EVC subgroup ($n = 8$), consisting of subjects with fewer than five EVC electrodes. As illustrated in Figure \ref{fig:spec_subgroup} and Table \ref{tab:freq}, subjects in the EVC subgroup with high-gamma inputs achieved notably higher median accuracy of 0.619, whereas the Non-EVC subgroup yielded a median accuracy of 0.423. This suggests that signals from the EVC region may contain richer neural presentations of the visual stimuli.

\begin{table*}[!htbp]
\centering
\caption{Group median $BA$ across frequency bands. Comparison across the full cohort, the EVC subgroup ($\geq 5$ electrodes in V1--V4), and the non-EVC subgroup ($< 5$ electrodes in V1--V4).}
\label{tab:freq}
\small
\begin{tabular}{lcccccc}
\toprule
\textbf{Group} & \textbf{Alpha} & \textbf{Beta} & \textbf{Low-gamma} & \textbf{Broadband-gamma} & \textbf{High-gamma} & \textbf{Broadband} \\ 
\midrule
Full Cohort ($n=17$) & 0.358 & 0.379 & 0.479 & 0.503 & 0.527 & 0.376 \\ \addlinespace
EVC Subgroup ($n=9$) & 0.454 & 0.421 & 0.568 & 0.607 & 0.619 & 0.416 \\ \addlinespace
Non-EVC subgroup ($n=8$) & 0.344& 0.322 & 0.373 & 0.365 & 0.423 & 0.343 \\
\bottomrule
\end{tabular}%
\end{table*}

An unanticipated result was observed in subject E04, who reached a decoding accuracy of 0.538 without any electrode covered in EVC. Therefore, we later extended our analysis to a wider range of cortical regions, described in Section \ref{subsec:regions}.

\begin{figure*}[htbp]
    \raggedright
    \begin{minipage}[t]{1.0\textwidth}
        \raggedright
        \textbf{A}\par\vspace{0.1cm}
        \includegraphics[width=\textwidth]{figs/attention_maps.png}
    \end{minipage}
    \par \vspace{-10pt}
    \begin{minipage}[t]{0.38\textwidth}
        \raggedright
        \textbf{B}\par\vspace{0.1cm}
        \includegraphics[width=\textwidth]{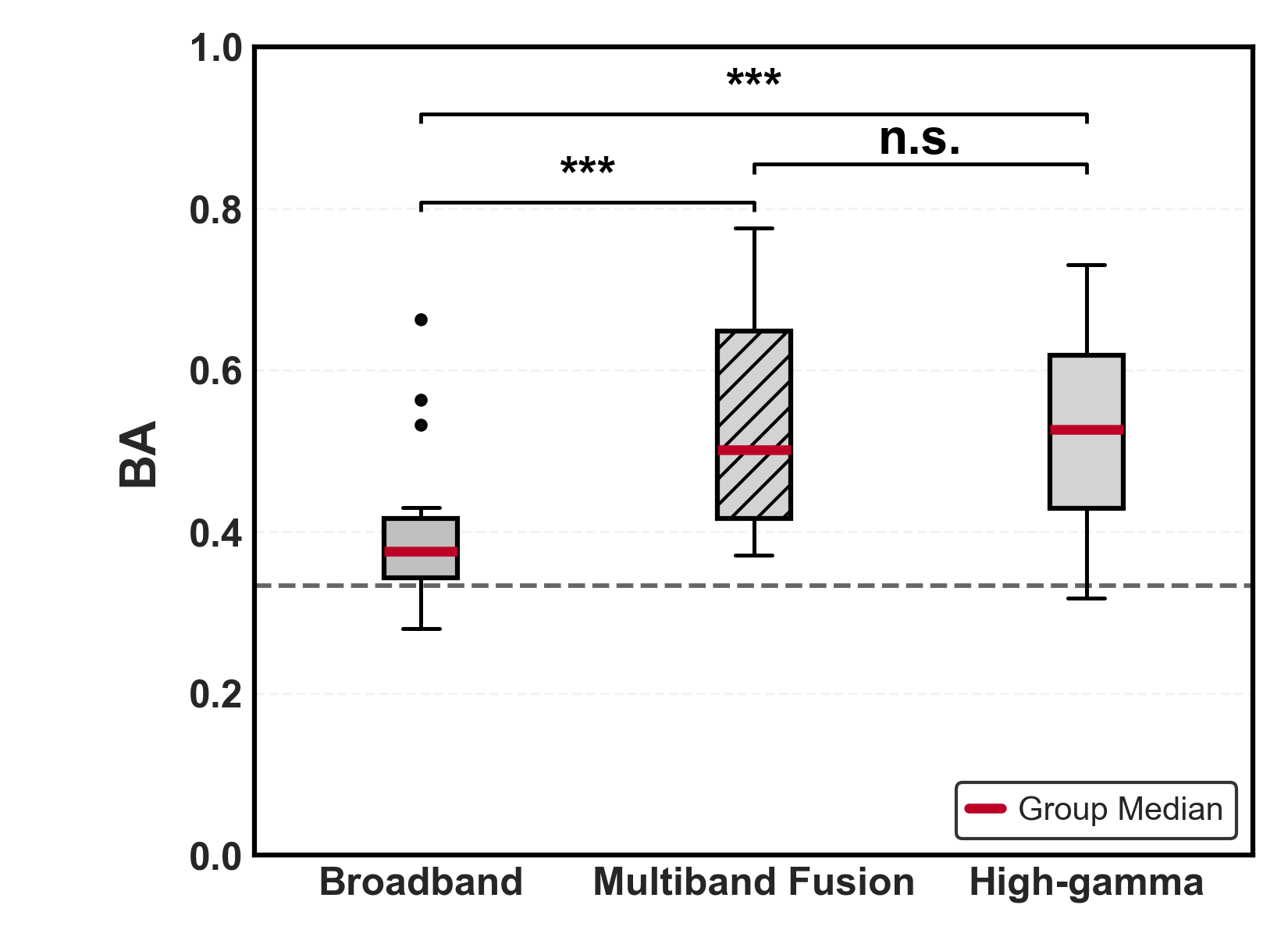}
    \end{minipage}
    \begin{minipage}[t]{0.41\textwidth}
        \raggedright
        \textbf{C}\par\vspace{0.1cm}
        \includegraphics[width=\textwidth]{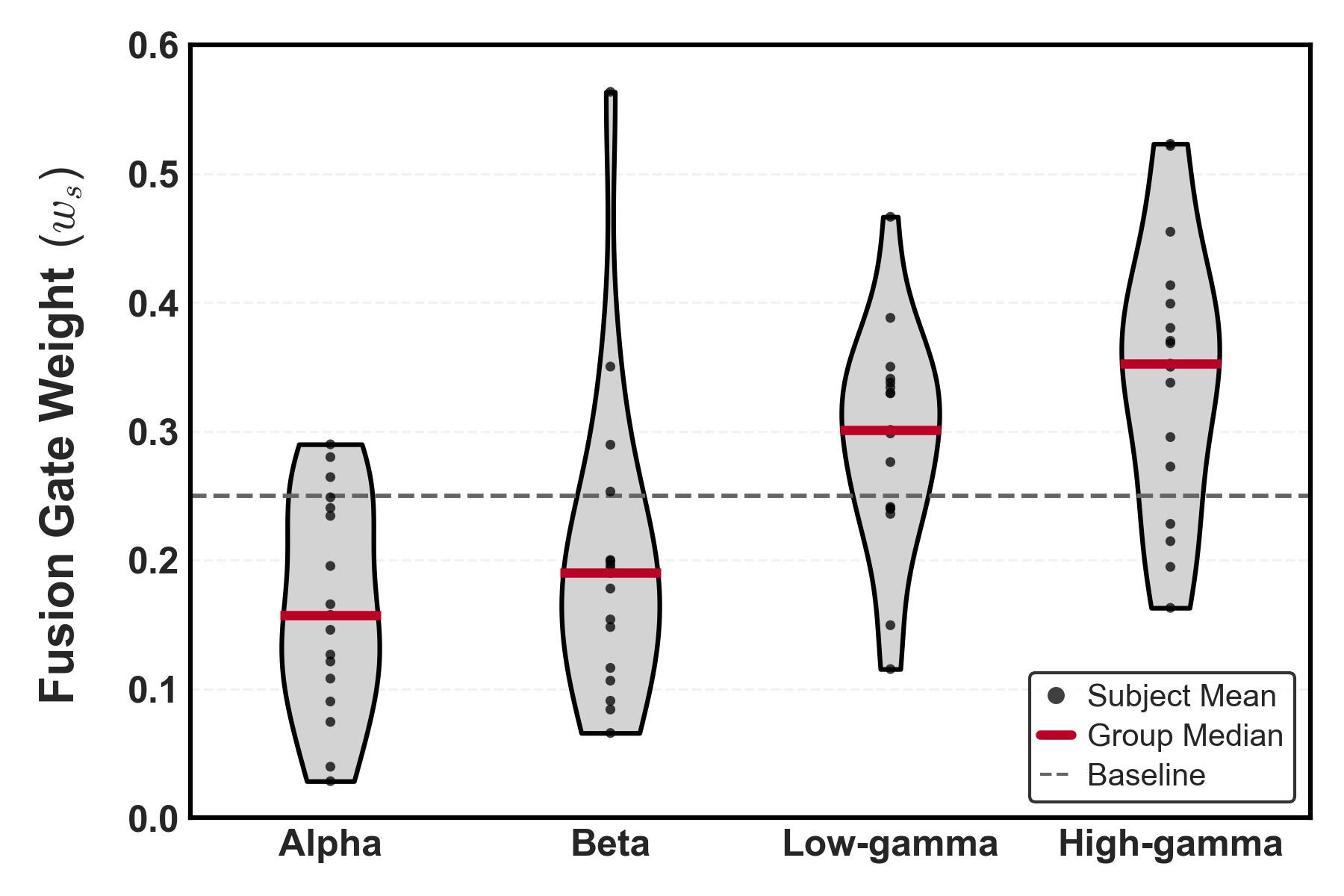}
    \end{minipage}
    \caption{
    \textbf{A}.~Last-layer self-attention heatmaps across frequency-band inputs for subject E02. Attention weights were averaged across heads and runs. Darker shade indicates large attention weight.
    \textbf{B}.~Decoding performance across broadband, multi-band fusion (spectrally disentangled), and high-gamma input representations ($n=17$). Asterisks indicate significance levels from pairwise Wilcoxon signed-rank tests after correction: $^{***}p<0.001$, $^{**}p<0.01$, $^{*}p<0.05$; n.s., not significant.
    \textbf{C}.~The distribution of learned fusion weights $w_s$ across different frequency-band inputs ($n=17$). The dashed baseline indicates equal band weighting ($w_s=0.25$).
    }
    \par \vspace{-10pt}
    \label{fig:attention_fusion}

    \subfiglabel{fig:att_heatmap}
    \subfiglabel{fig:fusion_compare}
    \subfiglabel{fig:gate_weights}

\end{figure*}

\subsection{Broadband Input Limits Transformer Decoding}
When using broadband inputs (3–250~Hz), the Transformer decoding performance only marginally exceeded chance level. This result is likely due to its self-attention mechanism. In this study, self-attention operates across the time dimension and treats individual time points as discrete temporal tokens. Without an explicit mechanism, the Transformer may struggle to disentangle distinct frequency band features distributed across tokens. This limitation is non-trivial as different frequency bands reflect different cortical information processing \citep{vonstein2000different}.

To test this rationale, we first visualised the last-layer attention weights across frequency band inputs. As illustrated in Figure \ref{fig:att_heatmap}, the narrow-band attention maps display off-diagonal high-attention stripes. The estimated spacings between adjacent stripes fall within the expected cycle duration of the corresponding frequency bands (see Table \ref{tab:spacing}). This suggests that the attention weights are sensitive to temporal structure in the input signals. However, in the broadband condition, the observed pattern cannot be aligned with a specific frequency component. The broadband attention map is less interpretable than other narrow band conditions. This suggests that the Transformer may have difficulty isolating frequency-specific temporal structure directly from broadband signals.

\begin{table*}[ht]
    \centering
    \caption{Approximate off-diagonal spacing in last-layer attention maps. Spacing was estimated from the attention map shown in Figure \ref{fig:att_heatmap}. }
    \small
    \begin{tabular}{l  c  c}
        \toprule
         \textbf{Band} & \textbf{Cycle Duration $T = 1/f$} (ms) & \textbf{Approx. Off-diagonal Spacing} (ms) \\
         \midrule
         Alpha (8-13Hz) & 76.9-125.0 & $\sim$100.0 \\
         \addlinespace
         Beta (13-30Hz) & 33.3-76.9 & $\sim$56.0\\
         \addlinespace
         Low Gamma (40-80Hz) & 12.5-25.0& $\sim$18.0\\
         \addlinespace
         High Gamma (80-150Hz) & 6.7-12.5& $\sim$10.0\\
        \bottomrule
    \end{tabular} \label{tab:spacing}
    \label{tab:placeholder}
\end{table*}

Motivated by this, we isolated the alpha, beta, low-gamma, and high-gamma bands from the broadband signals, and then fed each frequency-band input separately into the Transformer. In particular, each input was processed by the corresponding Transformer encoder from the preceding frequency-band experiment (see Section~\ref{subsec:cohort}), with its parameters frozen during subsequent training. This approach allows the model to extract neural representations from each band separately. We then projected these representations into a common latent space and generated intermediate embeddings $\mathbf{z}_s$. These embeddings $\mathbf{z}_s$ were then fused for classification as $\mathbf{z}_{\mathrm{fused}} =\sum_s w_s\mathbf{z}_s$ where $s \in \{ \text{alpha}, \text{beta}, \text{low-gamma}, \text{high-gamma} \}$ and $w_s$ denotes the relative weight. This projection and fusion process was entirely data driven and learned end-to-end. More details about this multi-band fusion model are in Figure \ref{fig:fusion_model} of the Supplementary Material. As a result, the cohort-level mean decoding accuracy of broadband inputs increased from 0.403 to 0.539 ($p<0.001$) (distribution shown in Figure \ref{fig:fusion_compare}). This divide-and-conquer approach allowed the Transformer to predict based on well-learned neural representations, thus overcoming the model limitation caused by spectral entanglement.

Moreover, as depicted in Figure \ref{fig:fusion_compare}, the median accuracy of high gamma was marginally higher than that of multi-band fusion ($p=1.0$). Surprisingly, providing additional frequency bands did not significantly improve model performance. The model may require a more sophisticated fusion strategy; we leave this for future work. In addition, Figure \ref{fig:gate_weights} presents the distribution of learned scalar weights for different frequency band inputs. These scalars were optimised through training. The result demonstrates that the model prioritised gamma features over alpha and beta during the decision-making process. Collectively, our findings further validate that the high-gamma input contains the most task-related information for the end-to-end decoding task.

\subsection{Contribution of Spectral Power and Temporal Phase}
We carried out a signal perturbation experiment to identify the ECoG features that contribute to decoding. In this experiment, we applied two types of signal perturbations to the ECoG input at the inference stage. In the constant-RMS perturbation, we replaced each channel waveform with a constant signal, where the amplitude was set to the root-mean-square of the original waveform. This preserved per-channel RMS power while destroying temporal and spectral structure. In the phase scrambling perturbation, each channel waveform was transformed into the Fourier domain. The magnitude spectrum was preserved, while the phase was randomised by drawing from a uniform distribution over $[0, 2\pi)$.

\begin{figure*}[ht]
    \raggedright
    \par \vspace{-10pt}
    \begin{minipage}[t]{0.30\textwidth}
        \raggedright
        \textbf{A}\par\vspace{0.1cm}
        \includegraphics[width=\textwidth]{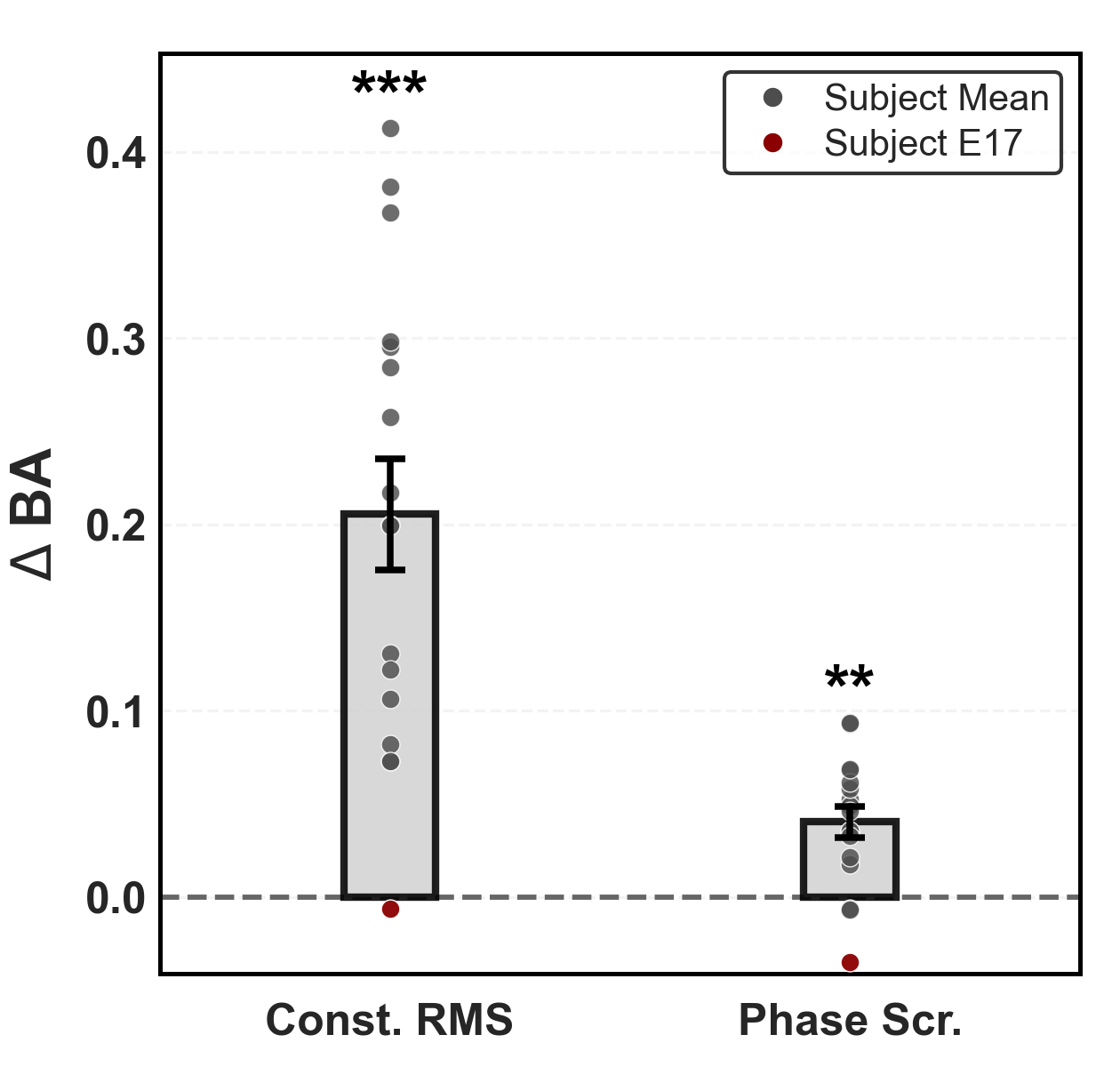}
    \end{minipage}
    \begin{minipage}[t]{0.68\textwidth}
        \raggedright
        \textbf{B}\par\vspace{0.1cm}
        \includegraphics[width=\textwidth]{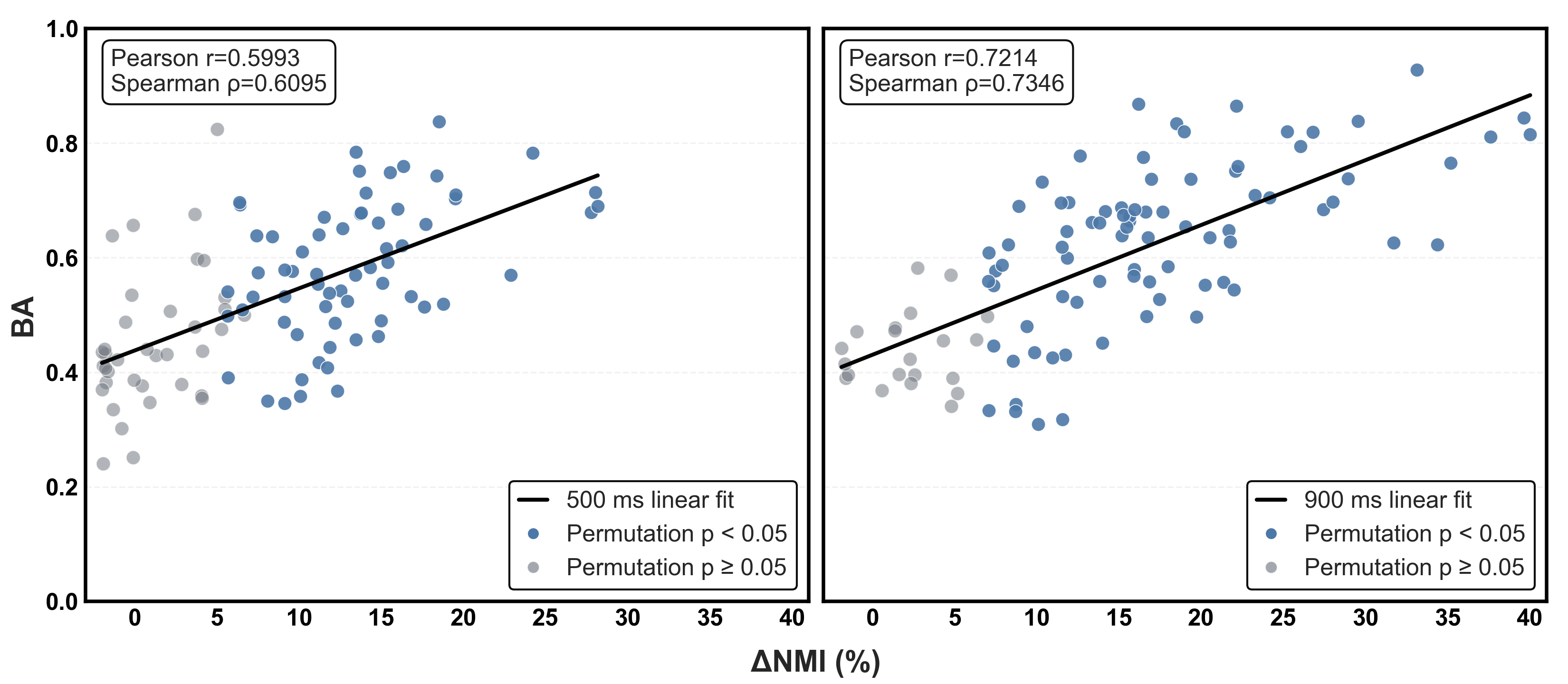}
    \end{minipage}

    \caption{
    \textbf{A}.~Effect of signal perturbations on high-gamma decoding performance. Bars show the mean $\Delta$BA relative to the unperturbed high-gamma baseline. Error bars indicate the standard error of the mean (SEM), dots show subject-level means, and asterisks indicate significance levels from Wilcoxon signed-rank tests against zero after correction: $^{***}p<0.001$, $^{**}p<0.01$, $^{*}p<0.05$.
    \textbf{B}.~Correlation between high-gamma spectral-power NMI and $BA$. Results are shown for the 500 ms and 900 ms post-stimulus windows. The 900 ms window was presented as it showed the optimal decoding performance in the post-stimulus window analysis. Dots show subject-fold results. Blue dots indicate significant MI by permutation test ($p<0.05$), and gray dots indicate non-significant MI ($p\geq0.05$).
    }
    \label{fig:perturb_mi}

    \subfiglabel{fig:drop}
    \subfiglabel{fig:mi}
\end{figure*}

As depicted in Figure \ref{fig:drop}, constant-RMS perturbation caused a significant decrease in accuracy across the cohort ($\text{mean}\ \Delta \approx 0.21$, $p<0.001$). Specifically, the majority of participants (except E17\footnote{E17 performed at near-chance levels, which may reflect poor data quality due to electrode-level or task-related factors (e.g., poor contact impedance or subject fatigue). This subject was, therefore, flagged for model-level analysis.}) showed accuracy declines ranging from 0.07 to 0.41. The phase scrambling perturbation presented marginal or even negative accuracy degradation ($\text{mean}\ \Delta \approx 0.04$), ranging from $-0.01$ to $0.09$ across subjects, excluding E17. These findings suggest that spectral power-related information of the high-gamma waveform was a major contributor, whereas phase structure made a smaller and secondary contribution. In other words, the Transformer primarily exploited the spectral power of the high-gamma inputs through the end-to-end learning framework.

To further examine whether high-gamma spectral power carried discriminative information for decoding, we estimated the mutual information (MI) between log-transformed high-gamma spectral power $\boldsymbol{Q}^{\mathrm{HG}} \in \mathbb{R}^{C}$ and visual-category labels $Y$, where $C$ denotes the number of selected channels. Specifically, we estimated MI between a continuous vector and a discrete variable using a k-nearest-neighbour MI estimator \citep{Ross2014} and computed the normalized MI as
\begin{equation}
\Delta \mathrm{NMI}
=100 \times
\frac{
\widehat{I}\left(\boldsymbol{Q}^{\mathrm{HG}};Y\right)
-\mathbb{E}\left[
\widehat{I}\left(\boldsymbol{Q}^{\mathrm{HG}};Y_{\mathrm{shuffle}}\right)
\right]
}{H(Y)}
\end{equation} 
where $\widehat{I}(\cdot)$ denotes the estimated MI, $H(Y)$ is the entropy of $Y$ and $\mathbb{E}\left[\widehat{I}(\boldsymbol{Q}^{\mathrm{HG}};Y_{\mathrm{shuffle}})\right]$ denotes the mean MI after label shuffling as a chance baseline. This baseline was estimated using 200 random label permutations within each training fold. The resulting $\Delta \mathrm{NMI}$ quantifies the percentage of label uncertainty $H(Y)$ explained by $\boldsymbol{Q}^{\mathrm{HG}}$. Figure \ref{fig:mi} shows that $\Delta \mathrm{NMI}$ was positively correlated with $BA$. This positive relationship indicates that high-gamma spectral power carried task-relevant information for visual category decoding.

\subsection{Impact of Post-stimulus Window Size}
In our prior analysis, a 500~ms post-stimulus window was used as the input epoch. This duration was selected to capture task-relevant responses to dynamic video stimuli \citep{kapeller2018realtime} while limiting noise from neural responses to subsequent, unrelated visual content. However, the high-gamma attention map (see Figure \ref{fig:att_heatmap}) revealed that global attention occurred near 500~ms. Hence, the 500~ms post-stimulus window might not be sufficient to include all task-relevant neural responses. For this reason, we investigated the impact of post-stimulus epoch window size on high-gamma decoding accuracy across the cohort.

\begin{figure*}[htbp]
    \raggedright
    \vspace{-10pt}
    \begin{minipage}[t]{0.43\textwidth}
        \raggedright
        \textbf{A}\par\vspace{0.1cm}
        \includegraphics[width=\textwidth]{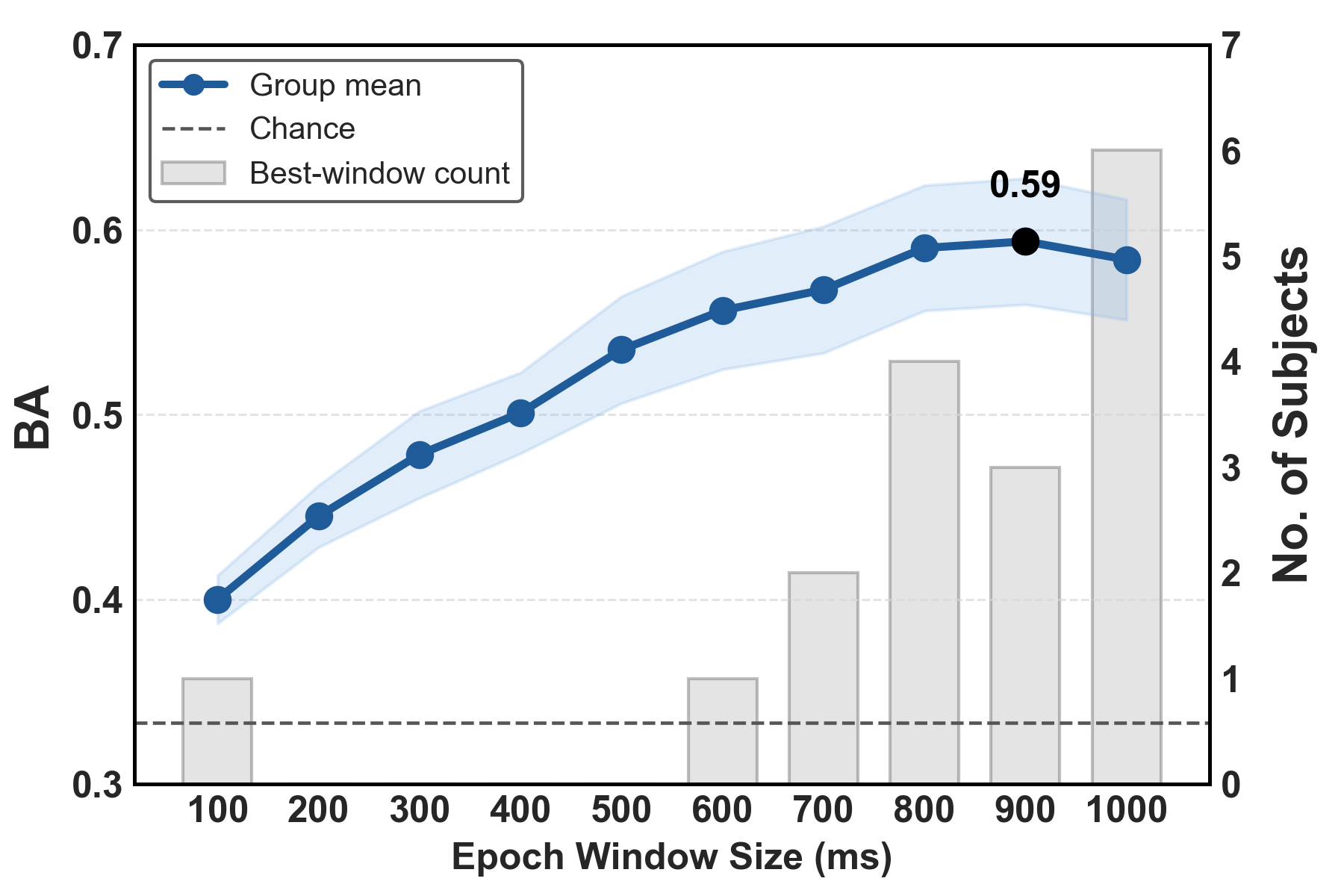}
    \end{minipage}
    \hspace{0.01\textwidth}
    \begin{minipage}[t]{0.55\textwidth}
        \raggedright
        \textbf{B}\par\vspace{0.1cm}
        \includegraphics[width=\textwidth]{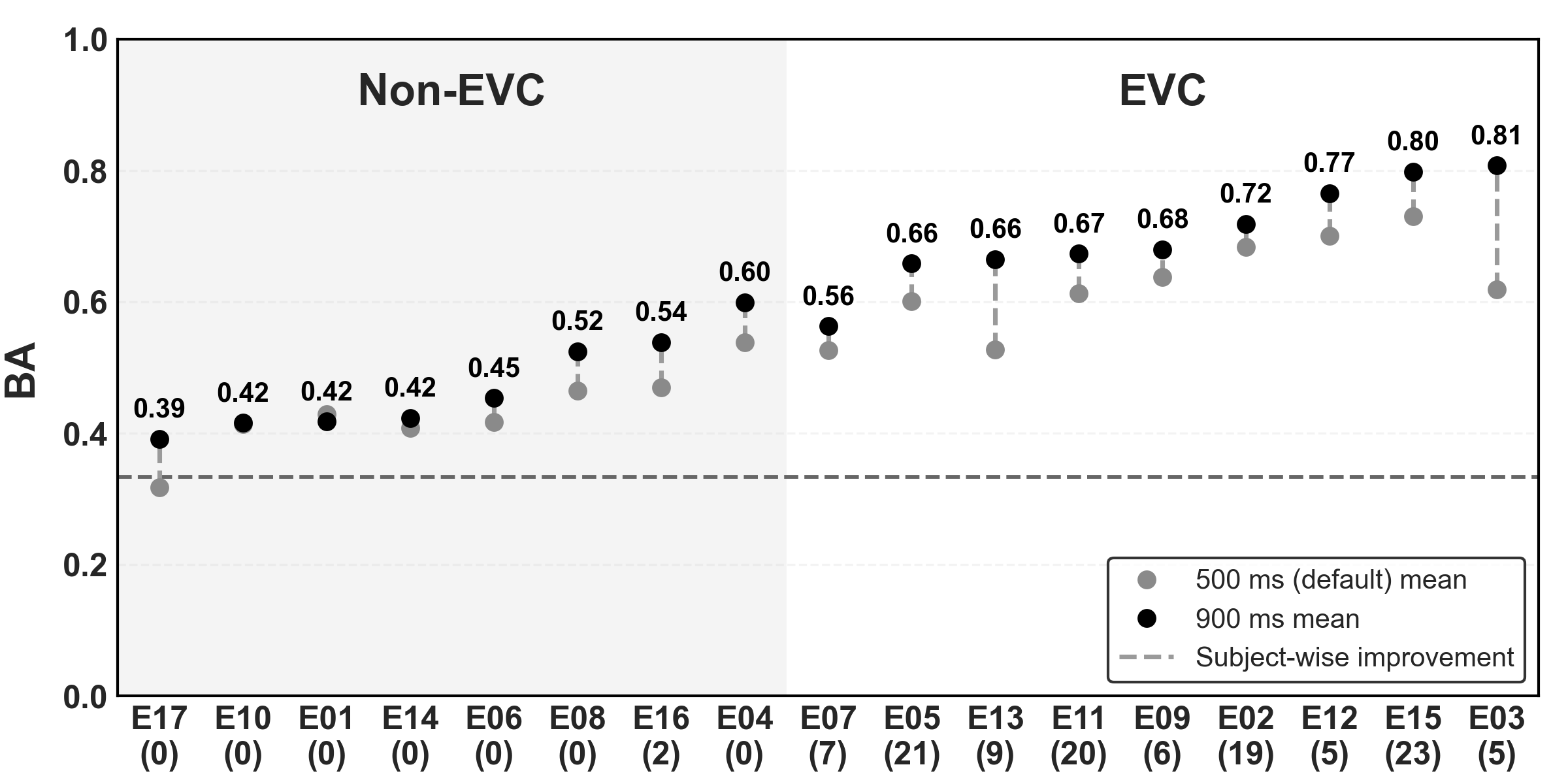}
    \end{minipage}
    \caption{
    \textbf{A}.~High-gamma decoding performance across post-stimulus epoch window size. The blue line shows group mean across subjects ($n=17$), and the shaded area indicates SEM. The black dot indicates the epoch window with the highest group mean. Gray bars show the distribution of optimal epoch window sizes across subjects.
    \textbf{B}.~Subject-wise comparison between 500~ms and 900~ms high-gamma decoding performance. Dashed gray lines indicate subject-wise performance change from 500 ms to 900 ms. Subjects are grouped as EVC or non-EVC according to whether they had ($\geq$ 5) electrodes across V1--V4. The numbers of subdural electrodes located in V1--V4 areas are shown in parentheses under subject IDs. Subjects are ranked within each group by mean BA.}
    \label{fig:window_900ms}

    \subfiglabel{fig:window}
    \subfiglabel{fig:900ms}

\end{figure*}

\begin{table*}[htbp]
\centering
\small
\caption{Effect of post-stimulus window size on high-gamma decoding performance. Mean $BA$ is reported across the full cohort ($n=17$), the EVC subgroup ($n=9$), and the non-EVC subgroup ($n=8$). \underline{Underlined} indicates the best performance within each group.}
\label{tab:window_size}
\begin{tabular}{lcccccccccc}
\toprule
\textbf{Group} & \textbf{100ms} & \textbf{200ms} & \textbf{300ms} & \textbf{400ms} & \textbf{500ms} & \textbf{600ms} & \textbf{700ms} & \textbf{800ms} & \textbf{900ms} & \textbf{1000ms} \\ 
\midrule
Full Cohort & 0.400 & 0.445 & 0.478 & 0.501 & 0.535 & 0.556 & 0.567 & 0.590 & \underline{0.594} & 0.584 \\ \addlinespace
EVC Subgroup  & 0.428 & 0.491 & 0.548 & 0.571 & 0.626 & 0.656 & 0.674 & 0.694 & \underline{0.703} & 0.668 \\ \addlinespace
Non-EVC Subgroup & 0.368 & 0.393 & 0.399 & 0.421 & 0.432 & 0.444 & 0.448 & 0.473 & 0.470 & \underline{0.489} \\
\bottomrule
\end{tabular}%
\end{table*}

Figure \ref{fig:window} and Table \ref{tab:window_size} show that decoding accuracy improved as the post-stimulus window was extended. Specifically, mean group accuracy increased from 0.400 at 100~ms to a peak of 0.594 at 900~ms\footnote{This analysis was limited to 1000 ms as the video stimuli were pre-segmented into 1~s intervals in the original dataset\citep{fukuma2022voluntary}. Window sizes beyond 1000~ms may introduce data leakage.}. This overall trend could be explained by the nature of the video stimuli. Dynamic video stimuli provide a richer semantic context compared to static images. Hence, the discriminative neural responses may not be solely concentrated in the early post-stimulus phase. Notably, the best window size varied across subjects. Fifteen participants achieved their best performance at longer window ranged from 700~ms to 1000~ms. Figure \ref{fig:900ms} illustrates subject-level performance using a 900~ms post-stimulus epoch window, compared with the 500~ms baseline. Sixteen subjects exhibited a clear improvement in $BA$ by at most 0.19, including all EVC subjects, when the epoch changed from 500~ms to 900~ms. Notably, the best individual performance was achieved by subject E03 using a 1000~ms post-stimulus window, reaching a mean $BA$ of 0.819. Furthermore, as shown in Figure \ref{fig:mi}, the correlation was stronger at 900~ms (Pearson $r=0.721$) than at 500~ms (Pearson $r=0.599$). This suggests that the 900~ms window carried additional task-relevant information from high-gamma spectral power compared to the 500~ms window.

\begin{figure*}
    \centering
    \includegraphics[width=1.0\textwidth]{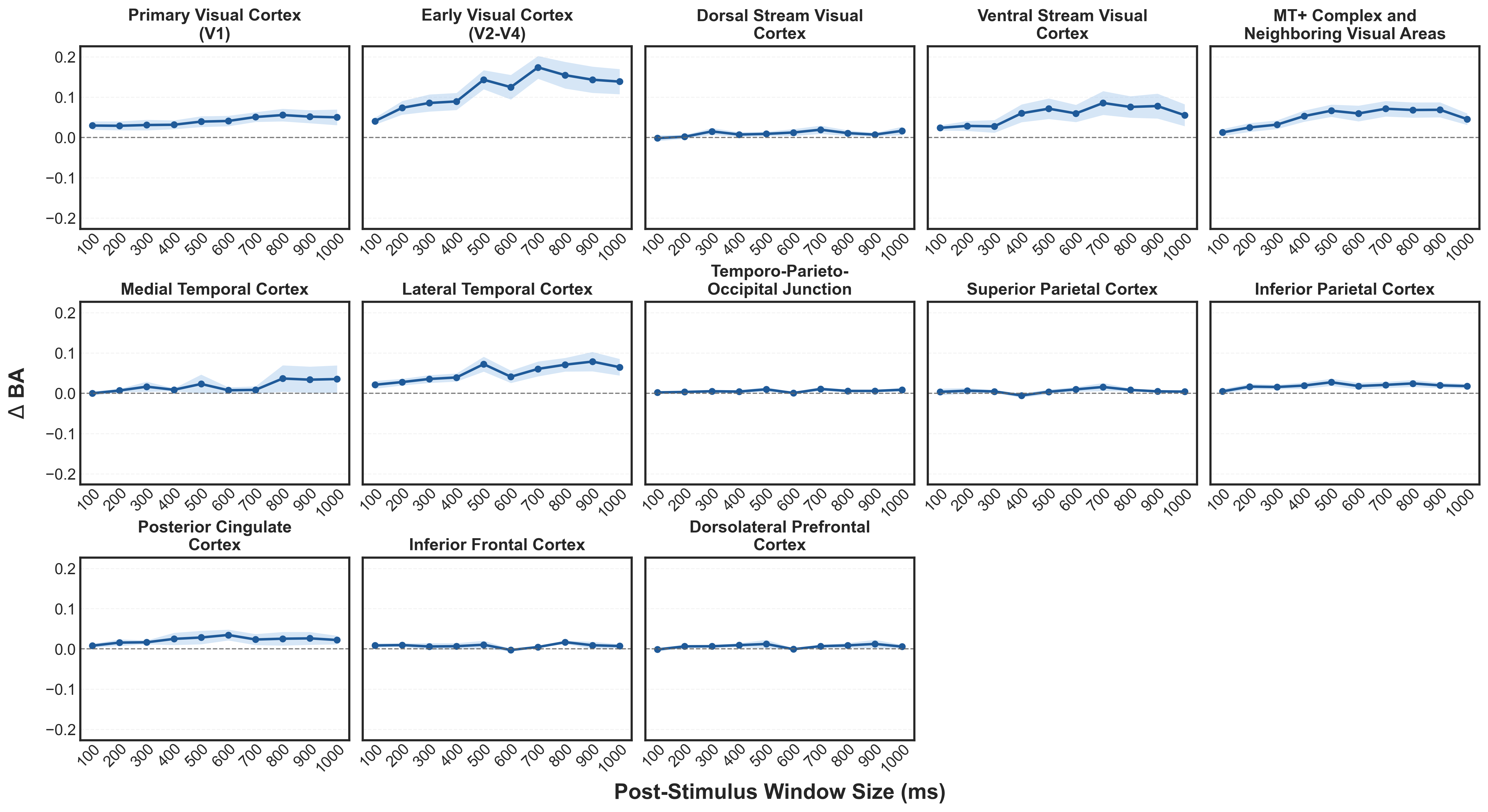}
 \caption{Mean region importance across post-stimulus epoch windows. Region importance was quantified as the mean $BA$ drop ($\Delta BA$) after negative replacement. Shaded areas indicate SEM across subjects ($n=17$).}
\label{fig:region}
\end{figure*}

\subsection{Cortical Region Importance} \label{subsec:regions}
To measure cortical region importance, we applied a permutation importance approach. For each cortical region of interest, we replaced its ECoG channels with the corresponding channels from a randomly selected negative sample. The negative example refers to the sample that is labeled as a different class. The region importance is defined as the performance decline in $BA$ (i.e., $\Delta BA$) compared to with unperturbed inputs. Figure \ref{fig:region} shows the results across cortical regions of interest. We found that early visual cortex (V2--V4), ventral stream visual cortex, MT+ complex with neighbouring visual areas, and lateral temporal cortex are important regions, as models exhibited an evident degradation in accuracy when replacing signals in these regions with false information. 

Early visual cortex ($n=10$) had dense electrode coverage, with an average of 8.5 electrodes per subject. Its importance score reached 0.143 at 500~ms, peaked at 0.174 at 700~ms, and maintained a mean score of 0.153 after 700~ms. This suggests that early visual cortex contributed task-relevant information over an extended post-stimulus period. The result also indicates that low-level visual features were highly discriminative for video stimulus categories in this study. However, the primary visual cortex had a lower contribution, where the mean importance score was 0.041 across window sizes. This may reflect the limited electrode coverage ($n=7$, Avg. electrodes = 4.6) or a limited contribution of neural responses within this region.

Lateral temporal cortex, which includes higher-order visual-associative regions, also participated substantially in the neural decoding task, maintaining a high importance score of 0.065 on average after 400~ms and showing a notable peak of 0.079 at 900~ms. This finding suggests that the signals from this higher-order visual processing region were used for decoding. Lateral temporal cortex ($n=17$) also had the broadest and consistent electrode coverage across participants with an average of 17 electrodes implanted. It reveals that this region is particularly relevant to clinical application, making it reliable and accessible for neural decoding.

The effects observed in the ventral stream visual cortex ($\text{mean}\ \Delta = 0.057$, $n=11$) and MT+ with neighbouring visual areas ($\text{mean}\ \Delta = 0.050$, $n=11$) were as expected, since ventral stream is usually associated with object and category recognition \citep{dicarlo2012visualobjectrecognition}, while MT+ is commonly associated with visual motion processing \citep{born2005structure}. Their contribution is thus expected for classifying faces, text, and landscapes from video. Notably, their importance scores started to increase when the post-stimulus window extended to 400~ms. This suggests that these regions may exhibit clear changes in neural responses after 300~ms post-stimulus. This finding broadly aligns with \citep{Kwon2021}, in which sustained broadband gamma (40-115~Hz) changes were prominently observed in visual and association cortical networks around 300-650~ms after stimulus onset. Thus, this observation suggests that the model also attended to later-stage activity, rather than relying only on early response.

We observed some effects in the inferior parietal cortex ($\text{mean}\ \Delta = 0.020$, $n=17$), showing that neural information from this region may still support the classification task but not substantially. However, other parietal and frontal regions demonstrated weaker effects. Importantly, while high importance scores indicate strong individual contribution, regions with near-zero importance scores should not be interpreted as having no contribution to the decoding process. Instead, their contribution may depend on interactions with other regions. The importance score in this study reflects only the contribution of each region in isolation and does not necessarily capture their full contribution to model inference.

Overall, the results are consistent with a hierarchical visual processing pattern. Specifically, the model used signals from early visual cortex, ventral stream visual cortex, MT+ complex with neighbouring visual areas, and lateral temporal cortex to support decision-making process for this visual semantic decoding task. The model's behavior can therefore be interpreted from a neurobiological perspective, which provides additional confidence in its reliability. 

\subsection{Related Work}
This study builds on the ECoG video-watching dataset collected by \citep{fukuma2022voluntary}. In that work, ECoG activity was mapped to a predefined semantic vector space to study whether imagery could shift the semantic representation derived from ECoG during conflicting visual stimulation. Here, we address a different problem: whether an end-to-end DL model can decode visual semantic categories directly from neural time-series, without manual feature engineering, under limited data conditions.

Recent work on visual semantic decoding has often aligned EEG signals with external image or multimodal representations from pretrained visual or vision-language models \citep{DBLP:conf/iclr/SongLLS0G24, DBLP:conf/nips/Li0LZL24}. Earlier related studies also used deep learning for visual category decoding, but they differ from our setting in important ways, such as using pretrained visual features to guide EEG decoding \citep{DBLP:conf/ijcai/JiaoYYLZS19}, converting EEG signals into spectrograms as the input to decoders \citep{ferrante2024decoding}, or repeated-trial EEG datasets with thousands of trials per subject \citep{kalafatovich2020decoding}. Most of these studies were based on EEG responses to still-image stimuli, where electrode layouts are generally more standardised across participants \citep{ferrante2024decoding, DBLP:conf/iclr/SongLLS0G24, DBLP:conf/nips/Li0LZL24, DBLP:conf/ijcai/JiaoYYLZS19, kalafatovich2020decoding}. Related ECoG studies have used temporal or spectral-temporal neural features to decode visual categories and semantic attributes \citep{majima2014decoding,rupp2017semantic}. In contrast, this study uses subject-specific ECoG responses to video stimuli, where electrode coverage is clinically determined, and trains a fully end-to-end model to predict semantic category labels directly, with fewer than 50 training samples per category. Moreover, we perform a series of analyses on the best-performing model to evaluate its reliability and interpretability.

\subsection{Limitation}
This study was conducted using a previously collected clinical ECoG dataset provided by \citep{fukuma2022voluntary} from participants with drug-resistant epilepsy. The findings may not fully generalise to broader populations and our interpretation was constrained by the available metadata. Since the subdural electrode placement was determined by clinical considerations, electrode coverage varied across regions and subjects. Because the visual semantic categories were selected from video stimuli, semantic category labels may be correlated with low-level visual features, such as contrast, spatial structure, motion, or scene composition. Therefore, the results should be interpreted as visual semantic decoding under video conditions.

Although we have considered and reported the number of subjects with electrode coverage in each region (see Table \ref{tab:ele_region_coverage} in the Supplementary Material) as well as the average number of electrodes within each region, the region importance analysis should be considered as a general, model-based estimate of cortical regional contribution. In this study, a fixed post-stimulus window was used to ensure consistent comparisons across frequency-band inputs, encoder architectures, and subjects. However, this fixed-window design may be less flexible for real-world scenarios, where neural response duration may vary across subjects and stimulus conditions. In addition, although a window-size analysis showed the 900 ms post-stimulus window as the group-level optimal window, the main model comparisons and frequency-band analyses were conducted using the predefined 500 ms baseline window. Hence, the 500 ms window may have underestimated the models' achievable performance in the earlier experiments. Furthermore, the attention and perturbation analyses are model-based interpretation. Therefore, the results from these analyses should not be considered as direct evidence of neural mechanisms or neurobiological processes.

\section{Conclusions}\label{sec5}
This study demonstrated that visual semantic categories in dynamic visual stimuli can be decoded end-to-end from ECoG time-series signals. We conducted a series of experiments and analyses in order to present a practical end-to-end approach for reliable visual decoding or, more broadly, for future BCI applications. The best-performing result was achieved by a Transformer-based encoder with mixup augmentation. The experimental results also showed that high-gamma activity was the most informative frequency-band input and that a 900~ms post-stimulus window provided the highest group-level mean balanced accuracy. As we believe that models should be evaluated beyond performance metrics, we further analyzed model behavior and demonstrated that it was interpretable across spectral, temporal, and cortical dimensions, and the interpretations were broadly consistent with domain knowledge. Together, these findings highlight the potential of data-efficient, interpretable, and reliable decoding systems to support practical BCI applications, particularly for individuals with severe neurological conditions.

\subsection*{Acknowledgements}
This research was funded by the Japan Science and Technology Agency (JST) Moonshot R\&D (JPMJMS2012).

\bibliographystyle{unsrtnat}
\bibliography{references}  

\newpage
\appendix

\section{Anatomical Regions}
\begin{table}[h]
    \centering
    \caption{Summary of Anatomical Region Selection}
    \small
    \begin{tabularx}{\textwidth}{l|X}
        \toprule
        & \textbf{Anatomical Regions} \\ 
        \midrule
        Included & Primary Visual Cortex, Early Visual Cortex, Dorsal Stream Visual Cortex, Ventral Stream Visual Cortex, MT+ Complex / Neighboring visual areas, Medial Temporal Cortex, Lateral Temporal Cortex, Temporo-Parieto-Occipital Junction, Superior Parietal Cortex, Inferior Parietal Cortex, Posterior Cingulate Cortex, Inferior Frontal Cortex, Dorsolateral Prefrontal Cortex \\
        \addlinespace
        Excluded & Somatosensory and Motor Cortex, Premotor Cortex, Posterior Opercular Cortex, Auditory Association Cortex, Orbital and Polar Frontal Cortex  \\
        \bottomrule     
    \end{tabularx}
    \label{tab:regions}
\end{table}

\begin{table}[htbp]
\centering
\setlength{\tabcolsep}{3pt}
\caption{Electrode counts by ROI and subject.}
\small
\label{tab:electrode-counts-by-roi-subject}
\resizebox{\textwidth}{!}{%
\begin{tabular}{l*{17}{r}}
\toprule
\textbf{ROI} & E01 & E02 & E03 & E04 & E05 & E06 & E07 & E08 & E09 & E10 & E11 & E12 & E13 & E14 & E15 & E16 & E17 \\
\midrule
Primary Visual Cortex (V1) & 0 & 6 & 0 & 0 & 5 & 0 & 1 & 0 & 2 & 0 & 8 & 0 & 2 & 0 & 8 & 0 & 0 \\
Early Visual Cortex (V2-V4) & 0 & 13 & 5 & 0 & 16 & 0 & 6 & 0 & 4 & 0 & 12 & 5 & 7 & 0 & 15 & 2 & 0 \\
Dorsal Stream Visual Cortex & 0 & 3 & 0 & 0 & 1 & 0 & 0 & 0 & 0 & 0 & 0 & 1 & 4 & 0 & 7 & 4 & 0 \\
Ventral Stream Visual Cortex & 0 & 3 & 6 & 0 & 3 & 0 & 2 & 1 & 4 & 0 & 11 & 10 & 0 & 0 & 10 & 1 & 1 \\
MT+ Complex and Neighboring Visual Areas & 0 & 5 & 4 & 0 & 7 & 0 & 9 & 0 & 6 & 0 & 14 & 0 & 2 & 4 & 4 & 7 & 2 \\
Medial Temporal Cortex & 7 & 0 & 0 & 1 & 5 & 4 & 0 & 1 & 2 & 0 & 0 & 1 & 0 & 5 & 1 & 0 & 4 \\
Lateral Temporal Cortex & 29 & 9 & 13 & 23 & 31 & 18 & 7 & 22 & 33 & 16 & 5 & 20 & 5 & 27 & 4 & 4 & 24 \\
Temporo-Parietal-Occipital Junction & 5 & 4 & 4 & 0 & 7 & 3 & 6 & 1 & 3 & 3 & 0 & 4 & 5 & 4 & 2 & 4 & 1 \\
Superior Parietal Cortex & 0 & 0 & 0 & 0 & 0 & 0 & 0 & 0 & 3 & 0 & 5 & 0 & 6 & 0 & 2 & 5 & 0 \\
Inferior Parietal Cortex & 3 & 17 & 13 & 2 & 1 & 7 & 5 & 3 & 25 & 6 & 10 & 4 & 16 & 6 & 11 & 22 & 6 \\
Posterior Cingulate Cortex & 0 & 2 & 0 & 0 & 2 & 0 & 0 & 0 & 0 & 0 & 11 & 2 & 0 & 0 & 2 & 1 & 0 \\
Inferior Frontal Cortex & 9 & 0 & 0 & 3 & 0 & 5 & 0 & 0 & 5 & 7 & 0 & 0 & 0 & 0 & 0 & 0 & 2 \\
DorsoLateral Prefrontal Cortex & 0 & 0 & 0 & 0 & 0 & 4 & 0 & 0 & 0 & 11 & 0 & 0 & 2 & 1 & 0 & 0 & 0 \\
\midrule
\textbf{Total electrodes} & \textbf{53} & \textbf{62} & \textbf{45} & \textbf{29} & \textbf{78} & \textbf{41} & \textbf{36} & \textbf{28} & \textbf{87} & \textbf{43} & \textbf{76} & \textbf{47} & \textbf{49} & \textbf{47} & \textbf{66} & \textbf{50} & \textbf{40} \\
\bottomrule
\end{tabular}%
}\label{tab:roi_ele}
\end{table}

\begin{table}[!hb]
\centering
\caption{Regional electrode coverage for region importance analysis across the full cohort. Electrodes identified as bad through manual inspection were excluded.}
\small
\begin{tabular}{lcc}
\toprule
\textbf{Region} & \textbf{No. Subjects} & \textbf{Avg. Electrodes / Subject} \\
\midrule
Primary Visual Cortex (V1) & 7 & 4.6 \\
Early Visual Cortex & 10 & 8.5 \\
Dorsal Stream Visual Cortex & 6 & 3.3 \\
Ventral Stream Visual Cortex & 11 & 4.7 \\
MT+ Complex \& Neighboring Visual Area & 11 & 5.8 \\
Medial Temporal Cortex (MTL / Hippocampus) & 10 & 3.1 \\
Lateral Temporal Cortex & 17 & 17.1 \\
Temporo-Parieto-Occipital Junction (TPO) & 15 & 3.7 \\
Superior Parietal Cortex & 5 & 4.2 \\
Inferior Parietal Cortex & 17 & 9.2 \\
Posterior Cingulate Cortex & 6 & 3.3 \\
Inferior Frontal Cortex & 6 & 5.2 \\
Dorsolateral Prefrontal Cortex & 4 & 4.5 \\
\bottomrule
\end{tabular}%
\label{tab:ele_region_coverage}
\end{table}

\clearpage

\section {Decoding Pipeline and Model Configuration} \label{ap:config}

\begin{table}[htbp]
    \centering
    \small
    \caption{Pooling and projection configuration for each encoder architecture based on preliminary ablation studies.}
    \begin{tabular}{l|c c}   
        \toprule
         \textbf{Encoder} &  \textbf{Pooling Strategy} & \textbf{Projection Head} \\ 
         \midrule
         ResNet & Temporal Attention & No \\
         LSNet & Temporal Attention & Yes \\
         TCN & Global Average & No \\
         LSTM & Global Average & Yes \\
         Transformer & Global Average & No \\
        \bottomrule     
    \end{tabular}
    \label{tab:config}
\end{table}

For each subject, hyperparameters were tuned before model training and evaluation, as subject-specific electrode coverage resulted in different input dimensions, which can affect model complexity and the optimal training configuration. 

\textbf{Hyper-parameter Tuning:} Bayesian optimization via Weights \& Biases (W\&B) sweep was employed for efficient hyper-parameter space search. Each sweep ran 50 trials. The hyperparameters include: (1) optimiser settings such as learning rate, weight decay and early stopping patience; (2) ANN architecture parameters including network depth, number of hidden units, kernel sizes and dilation factors; (3) regularisation parameters such as dropout rates and label smoothing; (4) training configuration including number of training epochs, batch size and channel normalisation strategy (z-score or median-IQR) and (5) strategy-specific parameters that must be defined prior to training. Importantly, this does not guarantee globally optimal settings for any architecture.

\textbf{Training Configuration:} The AdamW optimiser was used to optimise model with decoupled weight decay regularisation during training. We employed temporal 5-fold cross-validation with an 80/20 train-validation split within each fold. Please note that there was no overlap between the validation and test sets in any experimental run. To facilitate convergence, the learning rate was modulated by a ReduceLROnPlateau scheduler with a decay factor of 0.5, coupled with early stopping to mitigate overfitting. We applied channel dropout before the Transformer encoder during training, in which entire ECoG channels were randomly masked across the temporal window with probability $p$. The remaining channels were rescaled by $1/(1-p)$.

\textbf{Implementation Details:} Model training and evaluation were performed using Python 3.12.10 and PyTorch 2.4.1 with CUDA 12.1 on a workstation equipped with an NVIDIA L40S GPU (24 GB VRAM), 16 vCPUs, and 118 GB RAM.

\begin{table}[htbp]
\centering
\caption{Architecture of the Transformer-based model. Here, $C_s$ denotes the number of usable electrodes for subject $s$ after region selection and bad-channel exclusion, $T$ denotes the number of time samples, $d_s$ denotes the tuned hidden dimension, $L_s$ denotes the tuned number of Transformer blocks, and $H_s$ denotes the tuned number of attention heads. Channel dropout was applied during training.}
\label{tab:transformer}
\small
\begin{tabular}{lll}
\toprule
\textbf{Component} & \textbf{Description} & \textbf{Output Dimension} \\
\midrule
Input & Post-stimulus ECoG epoch & $C_s \times T$ \\
Channel scaling & Per-channel normalization & $C_s \times T$ \\
Channel projection & $1 \times 1$ convolution and batch normalization, $C_s \rightarrow d_s$ & $d_s \times T$ \\
Positional encoding & Sinusoidal positional encoding & $d_s \times T$ \\
Transformer block $\times L_s$ & Pre-normalized self-attention with $H_s$ heads, residual connection & $d_s \times T$ \\
& feed-forward network (linear $d_s \rightarrow 4d_s \rightarrow d_s$), dropout & $d_s \times T$ \\ 
Temporal pooling & Global average pooling over time & $d_s$ \\
Output layer & Linear classifier & 3 classes \\
\bottomrule
\end{tabular}%
\end{table}

\clearpage

\section{Multi-Band Fusion Model}\label{ap:fusion}
\begin{figure}[ht]
    \centering
    \includegraphics[width=\textwidth]{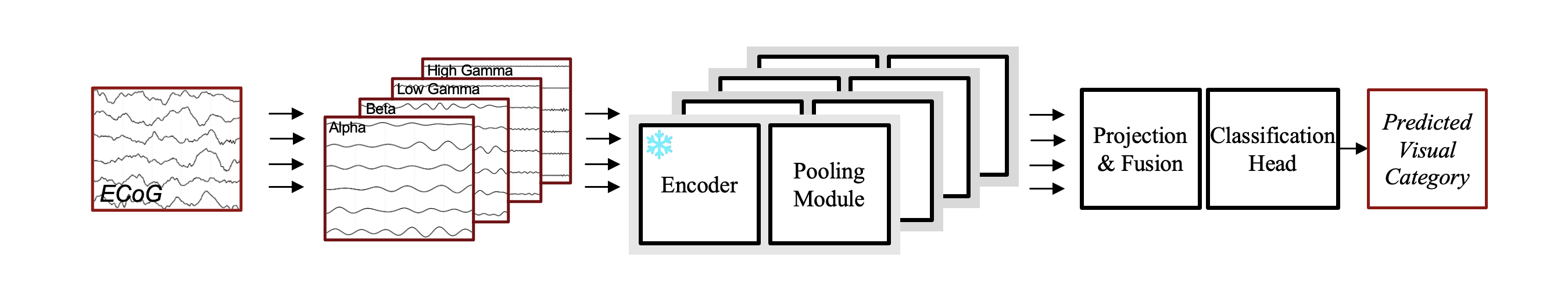}
    \caption{Architecture of multi-band fusion model. Each frequency band was processed by its corresponding pre-trained encoder obtained from previous experiment. These weights were frozen during training. The projection module consisted of a two-layer linear projection acting as a bottleneck. The fusion module computed relative weights $w_s$ to the embeddings $\mathbf{z}_s$ through scalar gates and softmax pooling. } \label{fig:fusion_model}
\end{figure}

\section{Additional Experimental Result}\label{ap:add_exp}
\begin{table}[ht]
\centering
\small
\caption{Subject-level high-gamma decoding performance across post-stimulus windows. For each subject, $BA$ was first averaged across 10 seeds within each fold, then reported as mean and standard deviation across folds. The final row reports the group mean and standard deviation across subject-level means. \underline{Underline} indicates the best performance.}
\label{tab:windows_subjects}
\begin{tabular}{lcccccccccccc}
\toprule
\multirow{2}{*}{\textbf{Subject}} 
& \multicolumn{2}{c}{\textbf{500 ms}} 
& \multicolumn{2}{c}{\textbf{600 ms}} 
& \multicolumn{2}{c}{\textbf{700 ms}} 
& \multicolumn{2}{c}{\textbf{800 ms}} 
& \multicolumn{2}{c}{\textbf{900 ms}} 
& \multicolumn{2}{c}{\textbf{1000 ms}} \\
\cmidrule(lr){2-3} \cmidrule(lr){4-5} \cmidrule(lr){6-7} 
\cmidrule(lr){8-9} \cmidrule(lr){10-11} \cmidrule(lr){12-13}
& Mean & SD
& Mean & SD
& Mean & SD
& Mean & SD
& Mean & SD 
& Mean & SD \\
\midrule
E01 & 0.429 & 0.11 & 0.472 & 0.13 & 0.405 & 0.06 & 0.439 & 0.10 & 0.418 & 0.09 & 0.433 & 0.08 \\
E02 & 0.683 & 0.04 & 0.718 & 0.02 & 0.677 & 0.03 & 0.750 & 0.07 & 0.718 & 0.04 & 0.687 & 0.05 \\
E03 & 0.619 & 0.05 & 0.613 & 0.04 & \underline{0.791} & 0.08 & 0.742 & 0.07 & \underline{0.808} & 0.07 & \underline{0.819} & 0.07 \\
E04 & 0.538 & 0.07 & 0.546 & 0.05 & 0.605 & 0.07 & 0.633 & 0.07 & 0.599 & 0.06 & 0.657 & 0.07 \\
E05 & 0.601 & 0.11 & 0.689 & 0.09 & 0.708 & 0.08 & 0.636 & 0.07 & 0.659 & 0.06 & 0.622 & 0.06 \\
E06 & 0.417 & 0.05 & 0.405 & 0.04 & 0.408 & 0.05 & 0.420 & 0.04 & 0.453 & 0.08 & 0.423 & 0.04 \\
E07 & 0.526 & 0.04 & 0.508 & 0.05 & 0.586 & 0.05 & 0.512 & 0.04 & 0.563 & 0.05 & 0.584 & 0.04 \\
E08 & 0.464 & 0.10 & 0.498 & 0.10 & 0.430 & 0.04 & 0.537 & 0.11 & 0.525 & 0.07 & 0.547 & 0.10 \\
E09 & 0.638 & 0.06 & 0.716 & 0.07 & 0.726 & 0.06 & 0.693 & 0.07 & 0.680 & 0.03 & 0.750 & 0.03 \\
E10 & 0.414 & 0.03 & 0.392 & 0.06 & 0.358 & 0.04 & 0.424 & 0.05 & 0.416 & 0.03 & 0.360 & 0.06 \\
E11 & 0.613 & 0.09 & 0.699 & 0.06 & 0.599 & 0.08 & 0.718 & 0.08 & 0.673 & 0.06 & 0.546 & 0.05 \\
E12 & 0.700 & 0.10 & \underline{0.732} & 0.11 & 0.729 & 0.10 & \underline{0.806} & 0.10 & 0.765 & 0.09 & 0.634 & 0.06 \\
E13 & 0.527 & 0.04 & 0.601 & 0.09 & 0.632 & 0.07 & 0.666 & 0.06 & 0.665 & 0.05 & 0.598 & 0.07 \\
E14 & 0.408 & 0.05 & 0.378 & 0.06 & 0.437 & 0.06 & 0.421 & 0.08 & 0.423 & 0.09 & 0.498 & 0.10 \\
E15 & \underline{0.730} & 0.08 & 0.629 & 0.06 & 0.620 & 0.05 & 0.720 & 0.04 & 0.798 & 0.10 & 0.770 & 0.05 \\
E16 & 0.470 & 0.07 & 0.527 & 0.10 & 0.583 & 0.10 & 0.532 & 0.09 & 0.538 & 0.07 & 0.615 & 0.07 \\
E17 & 0.318 & 0.07 & 0.333 & 0.03 & 0.355 & 0.05 & 0.380 & 0.04 & 0.391 & 0.05 & 0.381 & 0.04 \\
\midrule
Mean & 0.535 & 0.12 & 0.556 & 0.13 & 0.567 & 0.14 & 0.590 & 0.14 & 0.594 & 0.14 & 0.584 & 0.13 \\
\bottomrule
\end{tabular}%
\end{table}

\end{document}